\documentclass[letterpaper]{article} % DO NOT CHANGE THIS
\usepackage{aaai2026}  % DO NOT CHANGE THIS

\usepackage{helvet}  % DO NOT CHANGE THIS
\usepackage{courier}  % DO NOT CHANGE THIS
\usepackage[hyphens]{url}  % DO NOT CHANGE THIS
\usepackage{graphicx} % DO NOT CHANGE THIS
\usepackage{natbib}  % DO NOT CHANGE THIS AND DO NOT ADD ANY OPTIONS TO IT
\usepackage{caption} % DO NOT CHANGE THIS AND DO NOT ADD ANY OPTIONS TO IT
\usepackage{algorithm}
\usepackage{algorithmic}
\usepackage{xspace}
\usepackage{todonotes}
\usepackage{amsmath}
\usepackage{amsfonts}
\usepackage{booktabs}
\usepackage{subcaption} % For subfigures
\usepackage{textcomp}
\usepackage{lmodern}
\usepackage{mathptmx}

\usepackage{newfloat}
\usepackage{listings}
\DeclareCaptionStyle{ruled}{labelfont=normalfont,labelsep=colon,strut=off} % DO NOT CHANGE THIS
\floatstyle{ruled}
\newfloat{listing}{tb}{lst}{}
\floatname{listing}{Listing}
\newcommand{\mymodel}{\textsc{NodePro}\xspace}
\newcommand{\cora}{\textsc{Cora}\xspace}
\newcommand{\credit}{\textsc{Credit}\xspace}
\newcommand{\bitcoin}{\textsc{BitcoinAlpha}\xspace}
\newcommand{\fbkori}{\textsc{FB15k-237}\xspace}
\newcommand{\fbk}{\textsc{S-FB}\xspace}

\newcommand{\gcn}{\textsc{GCN}\xspace}
\newcommand{\gat}{\textsc{GAT}\xspace}
\newcommand{\gs}{\textsc{GraphSAGE}\xspace}
\newcommand{\mlp}{\textsc{MLP}\xspace}

\newcommand{\pt}{\textsc{Partial Train}\xspace}
\newcommand{\ft}{\textsc{Full Train}\xspace}

\newtheorem{RQ}{RQ}

\title{Draw a Portrait of Your Graph Data: An Instance-Level Profiling Framework for Graph-Structured Data}
\author{
    Tianqi Zhao\textsuperscript{\rm 1},
    Russa Biswas\textsuperscript{\rm 2},
    Megha Khosla\textsuperscript{\rm 1}
}
\affiliations{
    \textsuperscript{\rm 1}TU Delft\\
    \textsuperscript{\rm 2}Aalborg University\\
}

\usepackage{bibentry}
\usepackage{cleveref}
\begin{document}

\maketitle
% \todo[inline]{You are not really constructing a hypergraph. In a hypergraph an edge consists of more than 2 vertcies.}
%\todo[inline]{abstract too long}
\begin{abstract}
Graph machine learning models often achieve similar overall performance yet behave differently at the node level—failing on different subsets of nodes with varying reliability. Standard evaluation metrics such as accuracy obscure these fine-grained differences, making it difficult to diagnose when and where models fail. We introduce \mymodel, a node profiling framework that enables fine-grained diagnosis of model behavior by assigning interpretable profile scores to individual nodes. These scores combine data-centric signals—such as feature dissimilarity, label uncertainty, and structural ambiguity—with model-centric measures of prediction confidence and consistency during training. By aligning model behavior with these profiles, \mymodel reveals systematic differences between models, even when aggregate metrics are indistinguishable. We show that node profiles generalize to unseen nodes, supporting prediction reliability without ground-truth labels. Finally, we demonstrate the utility of \mymodel in identifying semantically inconsistent or corrupted nodes in a structured knowledge graph, illustrating its effectiveness in real-world settings.

%Overall, this work offers a step toward better understanding the relationship between graph data characteristics and model learning behavior, and introduces a practical approach for assessing prediction reliability on previously unseen nodes.

%and inject controlled errors such as random entities and flipped labels to simulate common sources of noise. We demonstrate that \mymodel effectively identifies these corrupted nodes. Overall, this work offers a step toward better understanding the relationship between graph data characteristics and model learning behavior, and introduces a practical approach for assessing prediction reliability on previously unseen nodes.
\end{abstract}

\section{Introduction}
\label{sec:intro}
Graph machine learning models have achieved remarkable success on node classification tasks. However, their performance is far from uniform across all nodes. Existing evaluation practices remain largely coarse-grained, relying on aggregate metrics such as accuracy or F1 score. While these metrics are valuable for benchmarking, they obscure important variation in how models behave at the node level. In particular, they fail to reveal why certain nodes are consistently harder to classify whether due to feature sparsity, structural ambiguity, or label noise.

Crucially, models trained on the same dataset and achieving similar overall accuracy may still generalize in fundamentally different ways—failing on different subsets of nodes or exhibiting varying levels of predictive uncertainty. Without a principled approach to capturing node-level learning difficulty and failure modes, we lack the tools to assess the trustworthiness of individual predictions.

Bridging this gap between high-level evaluation and fine-grained model behavior is essential for developing robust, interpretable, and trustworthy graph learning systems. This limitation motivates our central research question:
\textit{How can we characterize fine-grained differences in model behavior across individual nodes, beyond aggregate performance metrics?}

To address this question, we propose \mymodel, a node profiling framework designed to characterize individual nodes and explain model behavior at a fine-grained level. Rather than focusing on improving model performance through data augmentation or architectural changes, \mymodel seeks to uncover where models fail by identifying nodes that are consistently misclassified and analyzing the underlying factors that contribute to their difficulty. In doing so, our approach introduces a novel contribution to the growing field of data-centric graph machine learning \cite{zheng2023towards,Jin_2024}, which has traditionally emphasized improving datasets to enhance model performance. \mymodel instead shifts the emphasis toward understanding model behavior in relation to intrinsic node characteristics, offering a complementary path toward building more robust and interpretable graph learning systems.

\mymodel profiles nodes along two complementary dimensions: data-centric characteristics, which describe a node’s context in the input graph, and model-centric behavior, which captures prediction dynamics during training. The data-centric component constructs interpretable node profiles based on three key properties: (i) feature dissimilarity (how distinct a node’s features are relative to others in the same class), (ii) local label uncertainty (the diversity of class labels in its immediate neighborhood), and (iii) higher-order structural ambiguity (how consistently distant neighbors share the node’s label). Together, these properties summarize the intrinsic difficulty of a node from a data perspective.

To understand how models behave in relation to these profiles, \mymodel analyzes two additional signals: prediction consistency indicating how stable predictions are across training checkpoints, and prediction confidence which quantifies how far the predicted probability for the true label deviates from a random guess. By aligning these model-level signals with data-centric node profiles, \mymodel provides a powerful lens into model generalization and reliability.

We demonstrate that \mymodel not only explains node-level performance but also enables new capabilities such as detecting potentially incorrect predictions and flagging atypical or anomalous nodes, particularly in noisy or structurally complex graphs where traditional evaluation metrics fall short.

\section{Related Work}
%\todo[inline]{Check later and merge.}
\paragraph{Data-Centric Graph Machine Learning(DC-GML)}
Methods in DC-GML aim to improve model performance by focusing on the quality and manipulation of graph data itself, rather than solely on model architecture\cite{zheng2023towards,Jin_2024}. Recent research categorizes DC-GML into stages such data collection, improvement, and maintenance, emphasizing interventions on graph topology, features, and labels to combat noise, missing information, and mislabeling\cite{zheng2023towards, sun2024datacentricmachinelearningdirected, Jin_2024}.

Key strategies include augmenting node features to handle incompleteness, refining the labels, and using graph condensation or synthesis to improve the efficiency of the models\cite{zheng2023towards, Jin_2024, guo2023data, huang2024on}. These techniques have proven to be efficient to improve the performance of the existing SOTA models on tasks like GNN calibration, anomaly detection, and discovering hidden graph hierarchies. 

% Within the DC-GML context, graph anomaly detection, also known as graph out of distribution (OOD) detection, introduces taxonomies that classify nodes or entire graphs as hard versus easy, typical versus atypical, normal versus abnormal, high quality versus low quality, or in distribution (ID) versus out of distribution (OOD) \cite{guo2024data, 10.1145/3637528.3671929, gavrilev2023anomalydetectionnetworksscorebased}. These approaches are commonly applied to domain-specific graph datasets, such as those in bioinformatics, social networks, and financial systems. The primary objective is to perform data-centric manipulations that make the representations of OOD nodes/graphs more distinguishable from those of ID nodes/graphs. Detection strategies range from simple approaches, such as treating graphs with low local density as outliers \cite{10.1145/342009.335388}, to distance-based methods like $K$-nearest neighbors, where embeddings of graphs far from representation cluster centers are considered OOD \cite{sehwag2021ssd}. More sophisticated methods incorporate learnable parameters to increase the score gap between ID and OOD graphs during training, thereby enhancing detection performance \cite{10.1145/3580305.3599244}. while these work detect anomalies from the graph level, \cite{10.1145/3637528.3671929} proposed a diffusion-based graph generation method to synthesize training nodes, which can be promptly integrated to existing GNN models to boost their performances.

Within the DC-GML context, graph anomaly detection introduces taxonomies to classify nodes or entire graphs as hard/easy, typical/atypical, in distribution (ID) or OOD \cite{guo2024data, 10.1145/3637528.3671929, gavrilev2023anomalydetectionnetworksscorebased}. These methods are often applied to domain-specific datasets in bio-informatics, social networks, and finance. The goal is to make OOD representations more distinguishable from ID ones. Detection techniques include density-based heuristics \cite{10.1145/342009.335388}, distance-based methods such as $K$-nearest neighbors \cite{sehwag2021ssd}, and learnable scoring strategies that increase the separation between ID and OOD during training \cite{10.1145/3580305.3599244}. While most approaches focus on graph-level detection, \cite{10.1145/3637528.3671929} propose a diffusion-based graph generator to synthesize node-level training data, which can be directly integrated into existing GNN models to enhance performance.

Despite its promise, DC-GML still lacks frameworks for understanding how individual nodes affect model learning process. Most existing methods focus on global data properties or modifications, overlooking instance-level behavior.

\paragraph{Data Hardness in Data-Centric AI}
Data hardness in data-centric AI refers to identifying and managing difficult samples within datasets that impede model learning, particularly in image, tabular and visual tasks \cite{seedat2024dissectingsamplehardnessfinegrained, seedat2022dataiqcharacterizingsubgroupsheterogeneous}. These hard samples include mislabeled, ambiguous, or rare examples that challenge standard models. Addressing data hardness involves techniques like data augmentation to diversify training data and improve model robustness.

Hardness Characterization Methods are algorithmic tools used to detect such challenging samples. Recent research emphasizes evaluation of the data hardness quantitatively across benchmarks such as CIFAR and MNIST, aiming for reliable hardness detection and improved generalization of the models \cite{seedat2024dissectingsamplehardnessfinegrained}. Managing data hardness is critical for improving model accuracy, robustness, and trustworthiness in data-centric AI \cite{seedat2024dissectingsamplehardnessfinegrained, 10.1145/3571724}.

In node-level tasks on graphs, hardness is influenced not just by the features of a node, as in Euclidean data, but also by its surrounding topology, from local structure to higher-order neighborhoods. This added complexity calls for a holistic, graph-aware profiling approach that captures both node attributes and relational context.

\section{Our Proposed Framework}
\label{sec:my_model}
\label{sec:problem_formulation}

\begin{figure}
    \centering
    \includegraphics[width=1.0\linewidth]{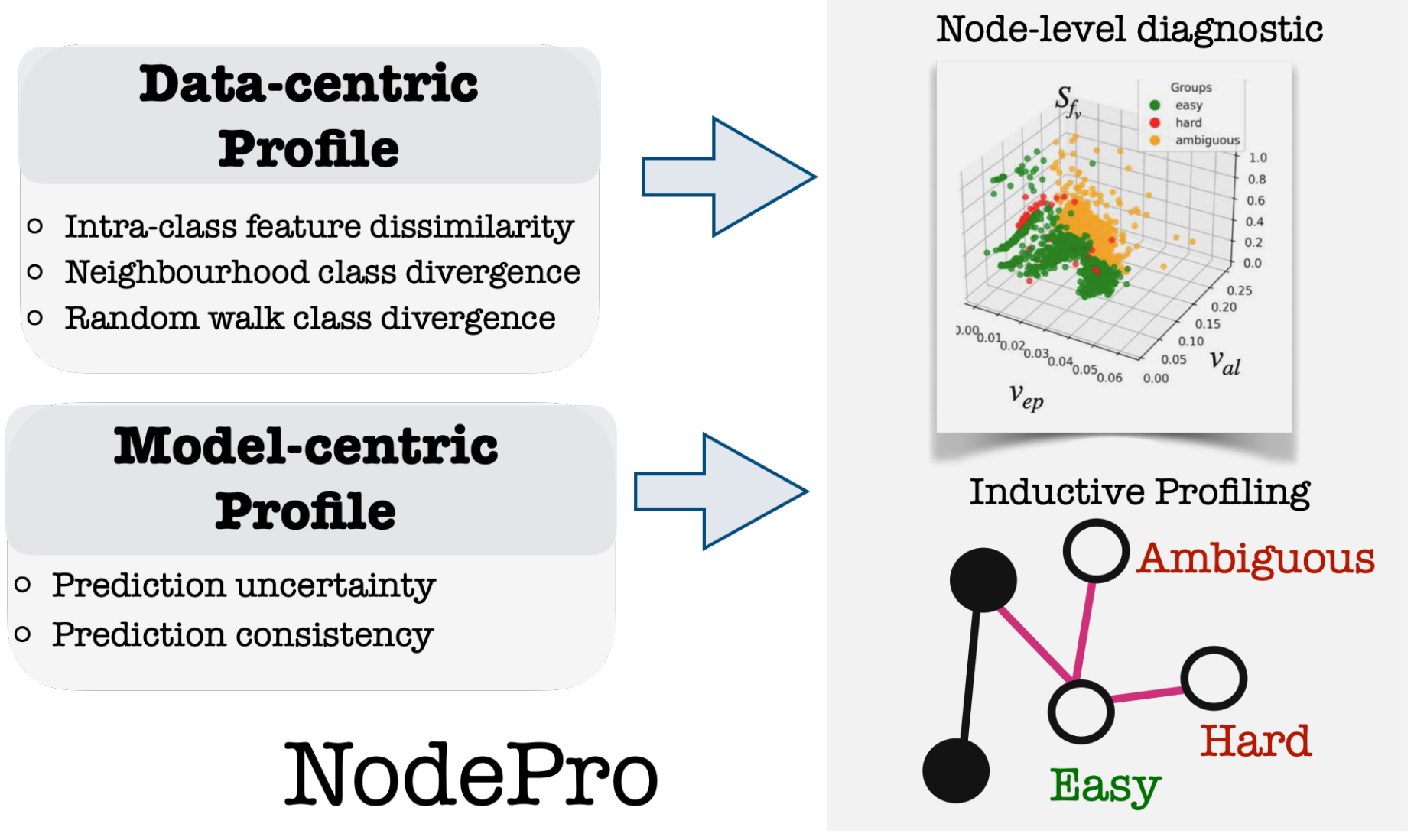}
    \caption{\mymodel combines data- and model-centric signals to diagnose node-level model behavior, enabling fine-grained model comparison, prediction reliability estimation, and error detection.}
    \label{fig:node_pro}
\end{figure}

We introduce \mymodel (see Figure \ref{fig:node_pro}), a node profiling framework designed to provide fine-grained insights into model behavior on graph-structured data. \mymodel consists of three main components: (i) data-centric node profiling that captures the intrinsic difficulty of nodes based on their features and structural context, (ii) model-centric node profiling module that analyzes prediction uncertainty and consistency across training, and (iii) an inductive profiling module that enables scoring of previously unseen nodes integrated into the graph. Together, these components allow \mymodel to characterize node-level learning behavior and identify failure modes beyond aggregate performance metrics. We begin by introducing the notations used throughout the paper and then describe the components of \mymodel. 

\paragraph{Notations} Given a graph $\mathcal{G} = \{\mathcal{V}, \mathcal{E}\}$, where $\mathcal{V} = \{v_1, \dots, v_n\}$ represents the node set and $\mathcal{E}$ represents the edge set, each node $v \in \mathcal{V}$ is associated with a feature vector $\mathbf{x}_v \in \mathbb{R}^d$, where $d$ is the feature dimension, and a label $y_v \in \{1, 2, \dots, C\}$, where $C$ is the total number of classes in the graph. We use $\mathcal{V}_c$ to denote all the nodes belong to class $c$, and $|\mathcal{V}_c|$ to indicate the size of class $c$. The one-hot encoded label vector of a node $v$ is denoted as $\mathbf{y}_v \in \{0, 1\}^C$.  In this work, we use the term "class" and "label" interchangeably.

\subsection{Data-Centric Node Profiling}
\label{sec:profiling_score}
This component computes interpretable scores that quantify the intrinsic difficulty of a node based on graph structure and input features. Specifically, it measures how different a node's features are from others in the same class, diversity of labels in the local neighborhood, and higher-order structural ambiguity of node. Each of these scores are detailed below.

\paragraph{Intra-class Feature Dissimilarity (\textsc{Icfd}).}
Intuitively, the nodes with input features that differ significantly from other nodes in the same class are harder to classify using feature-based models. These nodes lack the typical intra-class coherence and may therefore require additional contextual information (e.g., graph structure) for accurate prediction.

To quantify this, we introduce the \emph{intra-class feature dissimilarity} score, which captures the average cosine dissimilarity of a node $v$’s feature vector $\mathbf{x}_v$ from those of its class peers $\mathcal{V}_c$:

\begin{equation}
S_{f_v} = 1 - \frac{1}{|\mathcal{V}_c| - 1} \sum_{v' \in \mathcal{V}_c \setminus v} \frac{\mathbf{x}_v \cdot \mathbf{x}_{v'}}{\|\mathbf{x}_v\| \|\mathbf{x}_{v'}\|}
\end{equation}

This score serves as a lightweight, model-agnostic proxy for node difficulty: a high $S_{f_v}$ suggests that $v$ has atypical features relative to its class, and may challenge models that rely primarily on node attributes (e.g., MLPs). Conversely, a low score implies stronger intra-class alignment, making the node easier to classify from features alone. 

%As such, $S_{f_v}$ provides a principled way to profile node-level hardness from the perspective of feature informativeness.

\paragraph{{Neighborhood Class Divergence (\textsc{Ncd}).}}  We quantify the information captured in the one-hop neighborhood of a node $v$ by comparing the distribution of labels within this neighborhood with the average distribution of labels from the local neighborhoods of other nodes belonging to the same class. 
%\todo[inline]{too many unused notations , where is $$\mathcal{L}_v$$ used ? is it distribution of label set of node v? define $\mathcal{P}_v$ and $\mathcal{Q}_v$ before using them. Move the adaption to multi-label to apepndix }

This score is inspired by the notion of cross-class neighborhood similarity proposed in \cite{ma2023homophilynecessitygraphneural}, which extends beyond traditional homophily. As shown in \cite{ma2023homophilynecessitygraphneural}, effective class separation in GNNs does not require nodes of the same class to be directly connected. Instead, it is sufficient for their local neighborhoods to exhibit similar structural or label characteristics. Under this condition, models like GNNs can learn to embed such nodes into similar regions of the latent space, thereby facilitating accurate label prediction.

Specifically, we measure the difference of the label distribution in the direct neighborhood of node $v$ in class $c$ to the typical label distribution in the direct neighborhood of nodes in class $c$ using Kullback–Leibler divergence (KL-Divergence). Let $\mathcal{P}_v$ represent the normalized label distribution in the immediate neighborhood of node $v$, defined as:
$\mathcal{P}_{v}(c)= \frac{\sum_{u \in \mathcal{N}(v)} \mathbf{y}_u}{\sum_{u \in \mathcal{N}(v)} \sum_{c=1}^C \mathbf{y}_{u, c}}$, where $\mathcal{N}(v)$ denotes the set of one-hop neighbors of node $v$, and $\mathbf{y}_{u, c}$ denotes the $c$-th element in $\mathbf{y}_u \in \{0, 1\}^C$. Similarly, let $\mathcal{Q}_{y_v}$ denote the average normalized label distribution in the one-hop neighborhoods of all nodes in the same class as $v$, i.e., class $c$, given by: $\mathcal{Q}_{y_v} = \frac{1}{|\mathcal{N}_c|} \sum_{v \in \mathcal{V}_{c}} \sum_{u \in \mathcal{N}(v)} \mathbf{y}_u $, where $\mathcal{V}_{c}$ is the set of nodes with label $c$, and $\mathcal{N}_c$ is the union of one-hop neighborhoods for all nodes in $\mathcal{V}_{c}$.

\begin{equation}
S_{l_{v}} = \sum_{c\in \mathcal{C}} \left[ \log(\mathcal{P}_{v}(c) + \epsilon) - \log(\mathcal{Q}_{y_v}(c)) \right] \cdot (\mathcal{P}_{v}(c) + \epsilon)
\end{equation}

where $\epsilon$ is the smoothing factor, which takes the value of $10^{-10}$ in our implementation. We include a discussion of extending this formulation to multi-label graph dataset in Appendix.

\paragraph{\textbf{Random Walk Class Divergence (\textsc{Rwcd})}.} We introduce the Random Walk Class Divergence Score (RWCD) as a measure of how mixed the higher-order neighborhood of a node 
$v$ is with respect to class labels. The underlying hypothesis is that nodes surrounded by a higher proportion of different-class neighbors are harder to classify due to increased label ambiguity, typically occurring near class boundaries. 

We estimate the local class distribution around a node \( v \) by aggregating the labels encountered during multiple random walks. Specifically, let \( \mathcal{S} \) denote the \emph{multiset} of nodes visited across \( N \) random walks of length \( k \) starting from \( v \).  We compute the label count vector \(\mathbf{d}_w \in \mathbb{R}^C\)as
\(
\mathbf{d}_w = \sum_{j \in \mathcal{S}} \mathbf{y}_j ,
\)
which represents the frequency of each class observed in the random walk neighborhood of \( v \).
Finally, we define the RWCPS score denoted by $S_h$ as a measure of how mixed the neighborhood of node \( v \) is with respect to its true class label \( c \) as
\begin{align}
\label{eq:rwcps}
S_h = 1 - \frac{ \mathbf{d}_w[c] }{ \sum_{c' \in \mathcal{C}} \mathbf{d}_w[c'] },
\end{align}
where \( \mathbf{d}_w[c] \) denotes the number of nodes in \( \mathcal{S} \) belonging to the same class as \( v \).

\subsection{Model-Centric Node Profiling}
\label{sec:uncertainty_taxonomy}

To understand how models behave on individual nodes, \mymodel incorporates a model-centric profiling component that captures predictive dynamics during training. Specifically, we assess each node's \emph{prediction consistency} and \emph{prediction confidence} across model checkpoints, quantifying how stable and certain the model is about its predictions over time.
Following prior work \cite{seedat2022dataiqcharacterizingsubgroupsheterogeneous}, we use these uncertainty estimates to define a taxonomy of node difficulty, categorizing nodes as \textit{easy}, \textit{ambiguous}, or \textit{hard}. 

In particular, we consider a model $\mathcal{M}_{(\theta)}$ trained on a graph $\mathcal{G}$ by minimizing a supervised objective $\mathcal{L}$ over training data $\mathcal{D}_{\textit{train}}$. During training, we save a series of model checkpoints $\mathcal{E} = \{e_1, e_2, \dots, e_E\}$, each associated with a parameter configuration $\theta_e$.
At each checkpoint $e$, the model outputs a predicted class probability distribution over $C$ classes for every node $v \in \mathcal{V}$. Let $\mathcal{P}(v, \theta_e) \in [0,1]^C$ denote this predicted distribution at checkpoint $e$, and let $\mathcal{P}_c(v, \theta_e)$ denote the predicted probability for the true class label $c$ of node $v$. We define the \textit{mean predicted probability} for node $v$ over training as:
\[
\overline{\mathcal{P}}(v) = \frac{1}{E} \sum_{e \in \mathcal{E}} \mathcal{P}_c(v, \theta_e)
\]

We then compute two types of predictive uncertainty for each node: (i) \textbf{Epistemic uncertainty}, which captures the prediction variance across checkpoints, indicating model instability:
\begin{equation}
v_{\mathrm{ep}}(v) = \frac{1}{E} \sum_{e \in \mathcal{E}} \left[\mathcal{P}_c(v, \theta_e) - \overline{\mathcal{P}}(v)\right]^2
\label{eq:def_vep},
\end{equation}
and (ii) \textbf{Aleatoric uncertainty}, which measures the model's confidence in its prediction at each checkpoint:

\begin{equation}
    v_{\mathrm{al}}(v) = \frac{1}{E} \sum_{e \in \mathcal{E}} \mathcal{P}_c(v, \theta_e) \left(1 - \mathcal{P}_c(v, \theta_e)\right).
    \label{eq:def_val}
\end{equation}

These two quantities reflect complementary aspects of node-level difficulty: epistemic uncertainty signals inconsistency in learning, while aleatoric uncertainty reflects intrinsic ambiguity in the node’s label given its features and local structure.
Based on these quantities, we categorize each node $v \in \mathcal{V}$ into one of three classes: \textit{Easy}, \textit{Hard}, or \textit{Ambiguous}. This is formalized as follows:

\begin{equation}
g\left(v, \mathcal{G}\right) = \begin{cases}
\text{Easy} & \text{if } \overline{\mathcal{P}}(v) \geq C_{\text{up}} \;\wedge\; \\
             & v_{al}(v) < P_{50}\left[v_{al}(\mathcal{V})\right] \\[0.5em]
\text{Hard} & \text{if } \overline{\mathcal{P}}(v) \leq C_{\text{low}} \;\wedge\; \\
            & v_{al}(v) < P_{50}\left[v_{al}(\mathcal{V})\right] \\[0.5em]
\text{Ambiguous} & \text{otherwise}
\end{cases}
\label{eq:uncertainty_categorization}
\end{equation}

Here, $C_{\text{up}}$ and $C_{\text{low}}$ are confidence thresholds, and $P_{50}[\cdot]$ denotes the empirical median of a distribution. 

% \todo[inline]{I am not sure if it is a great difference to point out }
% Note that different from \cite{seedat2022dataiqcharacterizingsubgroupsheterogeneous}, we categorize all the nodes instead of just training nodes in graph $\mathcal{G}$, and use this categorization as ground-truth for the further analysis.

\subsection{Inductive Profiling of Unseen Nodes}  
\label{sec:Inductive_Profiling}

\mymodel enables profiling of previously unseen nodes introduced into the existing graph $\mathcal{G} = (\mathcal{V}, \mathcal{E})$. Given a new node $v_{\text{new}}$ with feature vector $\mathbf{x}_{\text{new}}$ and incident edges $\mathcal{E}_{\text{new}}$, we construct an augmented graph $\mathcal{G}' = (\mathcal{V} \cup \{v_{\text{new}}\}, \mathcal{E} \cup \mathcal{E}_{\text{new}})$. Given a model $\mathcal{M}_{(\theta)}$ trained on $\mathcal{G}$ with saved checkpoints $\mathcal{E} = \{e_1, \dots, e_E\}$ we use the final checkpoint $e_E$, to compute the representation of $v_{\text{new}}$ via a forward pass on its local subgraph consisting of $L$-hop neighborhood for an $L$-layer GNN:
$\hat{\mathbf{z}}_{v_{\text{new}}}^{(e_E)} = \mathcal{M}_{(\theta_{e_E})}(\mathcal{G}_{\text{sub}}).
$
We then perform $K$-nearest neighbor search in the representation space to find $\mathcal{N}_K(v_{\text{new}})$ from existing nodes. \mymodel predicts a difficulty category (easy, hard, or ambiguous) via majority vote over this neighborhood:
\begin{equation}
    \hat{P}_{v_\text{new}} = \text{MajorityVote} \left( \{\text{Category}_u \mid u \in \mathcal{N}_K(v_{\text{new}})\} \right)
\end{equation}

To evaluate profiling accuracy, we derive the ground-truth difficulty category for $v_{\text{new}}$ by computing prediction consistency and confidence across checkpoints (Equations~\ref{eq:def_vep}–\ref{eq:uncertainty_categorization}). Categorization accuracy is then:
\begin{equation}
    \text{Acc} = \frac{1}{|\mathcal{V}_{\text{new}}|} \sum_{v \in \mathcal{V}_{\text{new}}} \mathbb{I}(\hat{P}_v = P_v)
\end{equation}

Nodes classified as \emph{easy} tend to receive confident and consistent predictions, while \emph{hard} or \emph{ambiguous} nodes exhibit instability during training, indicating potentially unreliable predictions.

\section{Research Questions and Experimental Setup}
\label{sec:experiment}
To validate and examine the broader implications of \mymodel, which has been specifically designed to provide fine-grained characterizations of model behavior at the node level (thereby addressing the central research question posed in Section Introduction), we explore the following research questions:
\begin{RQ}
 How can \mymodel be used to uncover fine-grained differences in model behavior at the node level between models with similar overall accuracy?
\end{RQ}

\begin{RQ}
    To what extent can \mymodel accurately predict the reliability of model predictions on previously unseen nodes?
\end{RQ}

\begin{RQ}
    How effectively can \mymodel identify semantically inconsistent or erroneous instances in knowledge-driven graph settings?
\end{RQ}
\subsection{Datasets}We demonstrate \mymodel on three diverse real-world datasets: (i) \textbf{\cora}~\cite{DBLP:journals/corr/YangCS16}, a citation network where nodes are research papers with binary word features and class labels based on topic categories; (ii) \textbf{\credit}~\cite{DBLP:journals/corr/abs-2102-13186}, a financial graph with customer nodes and attributes related to credit risk, where we use a preprocessed subset of $3000$ nodes following~\cite{Olatunji_2023}; and (iii) \textbf{\bitcoin}~\cite{7837846}, a signed, directed trust network of trading accounts, where node features are derived from trust scores and structural statistics.
In addition, to demonstrate the ability of \mymodel to identify atypical patterns and semantic inconsistencies beyond what standard benchmarks reveal we construct a node classification task from the \fbkori knowledge graph dataset~\cite{toutanova2015observed} as described below and further detailed in the Appendix. 
%Our goal is to evaluate whether \mymodel can distinguish nodes encoding correct knowledge from those containing semantic errors or injected label noise.

\paragraph{Knowledge Graph \fbk.}
Starting from the training subset $\mathcal{G}_{\textit{train}}$ of \fbkori \cite{toutanova-chen-2015-observed}, we create a new graph $\mathcal{G}_{\text{s}} = (\mathcal{V}_s, \mathcal{E}_s)$ where each node represents a triplet $U = \langle u, r, v \rangle$. Two nodes are connected if they share an entity. Textual representations are generated using the \texttt{intfloat/multilingual-e5-large} model by concatenating the descriptions of the entities and relation. We pose a binary classification task: nodes derived from valid triplets are labeled as \emph{true} ($1$), and corrupted variants—constructed by replacing one entity in $1\%$ of triplets with a random entity—are labeled as \emph{false} ($0$). To further simulate label noise, we randomly flip the labels of $30\%$ of nodes. This setup allows us to test whether \mymodel can flag semantically incorrect or mislabeled nodes as hard or ambiguous, thereby demonstrating its utility for error detection in knowledge-driven settings.

We further summarize the characteristics of the datasets in the following Table \ref{tab:datasets_description}.

\begin{table}[h!]
\small
\centering
\setlength{\tabcolsep}{3pt}
\begin{tabular}{lcccccc}
\toprule

\textbf{Dataset} & \textbf{\#Nodes} & \textbf{\#Edges}  & \textbf{\#Features} & \textbf{\#Classes}  \\
\midrule
\cora           & $2,708$   & $5,429$     & $1,433$   & $7$    \\
\credit        & $3,000$   & $28,854$    & $13$      & $2$     \\
\bitcoin  & $3,783$   & $28,248$   & $8$      & $2$    \\
\fbk    &$272,115$ &$996,010,50$ &$1,024$ &$2$ \\ 

\bottomrule
\end{tabular}
\caption{Summary of Graph Dataset Characteristics}
\label{tab:datasets_description}
\end{table}
%\todo[inline]{compare the behavior of models with the same acc: gat and gs for cora in \pt, \gcn and \gs for \credit in \pt, \gcn and \gs for bitcoin in \pt, cora gcn obtains the highest acc in \pt, mlp for \bitcoin in pt;

%gs and mlp on cora in \ft, \gcn and \gat for credit, \gat and \mlp for bitcoin \ft}
\subsubsection{Models and training set up}
We consider three widely used graph neural networks: \gcn \cite{kipf2017semisupervisedclassificationgraphconvolutional}, \gat \cite{veličković2018graphattentionnetworks}, and \gs \cite{hamilton2018inductiverepresentationlearninglarge}, along with a simple multilayer perceptron (\mlp) as a baseline. In the standard training setup, we used the built-in split in \cora and \credit and randomly split the nodes in \bitcoin into $60\%$ for training, $20\%$ for validation, and $20\%$ for testing. We denote this setting as \pt in the following sections. Additionally, to assess the upper bound of model performances, we adopt a second setup in which all nodes in the graph are used simultaneously for training, validation, and testing. We denote this setting as \ft. The training process in these two setups are combined with early-stopping with patience of $100$ epochs.

\begin{table}[!ht]
    \small
    \centering
    \begin{tabular}{lccc}
    \toprule
            &\cora  &\credit  &\bitcoin \\\midrule
         \gcn  &$0.801 / 0.981$  &$0.613 / 0.661$  &$0.850 / 0.850$ \\
         \gat  &$0.780 / 0.997$  &$0.594 / 0.641$  &$0.594 / 0.732$ \\
         \gs  &$0.780 / 1.000$  &$0.614 / 0.690$  &$0.810 / 0.850$ \\
         \mlp  &$0.569 / 1.000$  &$0.623 / 0.680$  &$0.738 / 0.740$ \\
    \bottomrule
    \end{tabular}
    \caption{Prediction Accuracy in \pt / \ft setups. The full table with standard deviation on three random splits is summarized in Appendix.}
    \label{tab:nc_acc_pt_ft}
\end{table}

\section{Results}
% Due to page limits, we present a summary and analysis of the results in the form of answers to the previously stated research questions. Detailed analyses of each model’s behavior across different setups and datasets are provided in the Appendix.
We now demonstrate how \mymodel can be used to compare different models in the behavior at the node level \textbf{(RQ1)}, estimation of prediction reliability \textbf{(RQ2)} and  detection of semantic errors \textbf{(RQ3)}.
\subsection{Differences in Model-Behavior (RQ1)}
As shown in Table~\ref{tab:nc_acc_pt_ft}, different models often attain comparable overall accuracy on the same dataset. However, this raises a critical question: do these models behave similarly at the node level despite matching aggregate performance? In this section, we  leveraging \mymodel to uncover fine-grained differences in model behavior, enabling us to move beyond coarse evaluation metrics and diagnose how and on which nodes models diverge in their learning and generalization patterns.

\begin{figure*}[!ht]
  
    \begin{subfigure}[b]{0.49\linewidth}
        \centering
        \includegraphics[width=\linewidth]{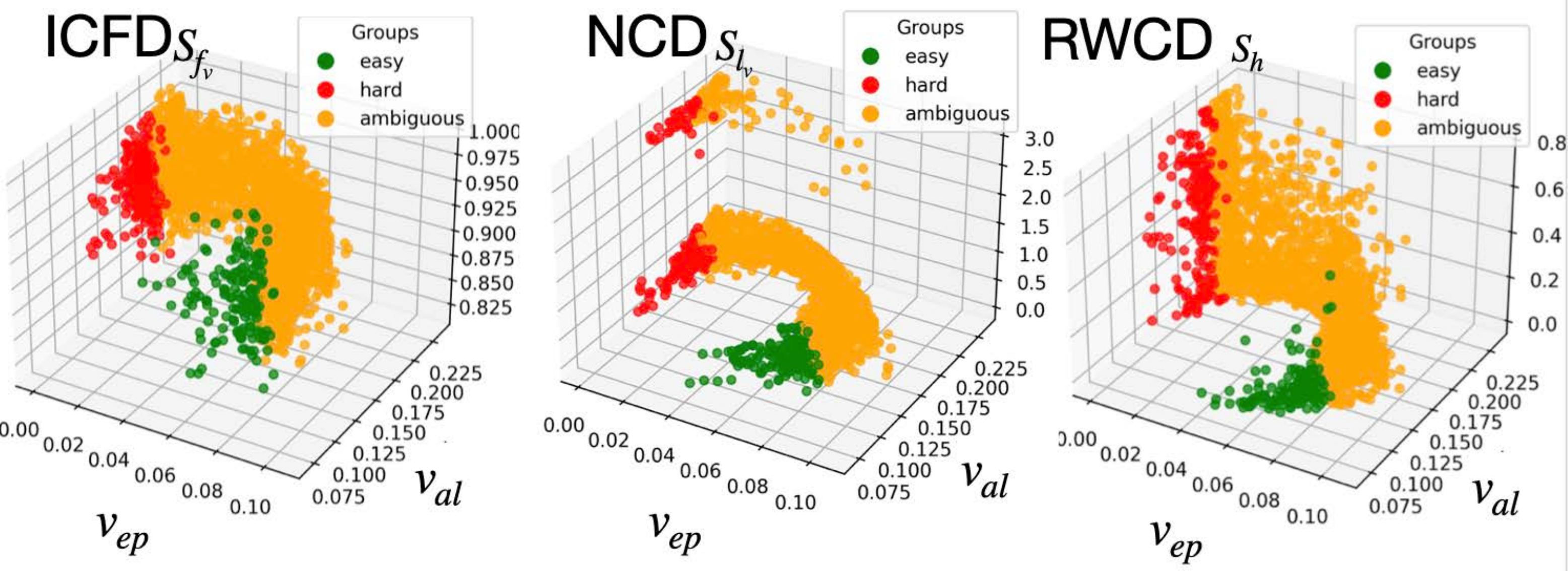}
        \caption{\textsc{GAT} on \textsc{Cora}.}
        \label{fig:correlation_scatter_partial_train_gat_cora}
    \end{subfigure}
    \begin{subfigure}[b]{0.49\linewidth}
        \centering
        \includegraphics[width=\linewidth]{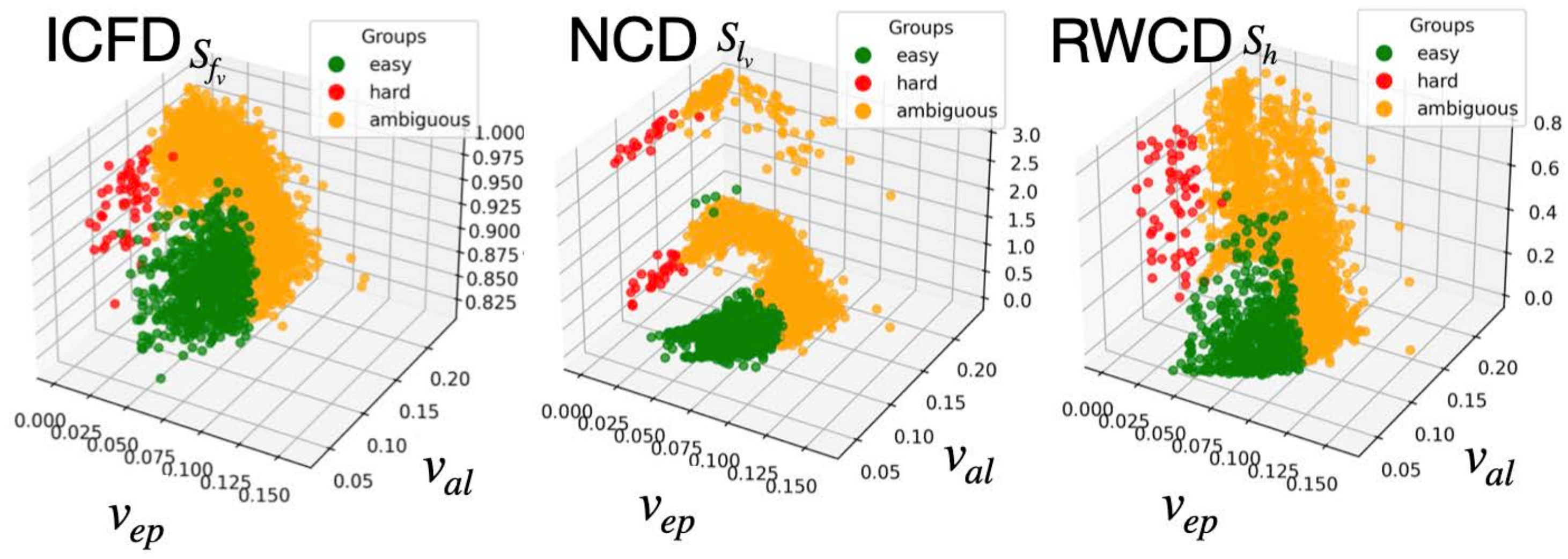}
        \caption{\textsc{GraphSAGE} on \textsc{Cora}.}
        \label{fig:correlation_scatter_partial_train_gs_cora}
    \end{subfigure}
        \caption{Diagnosing \gat and \gs when trained with built in splits of \cora. }
    \label{fig:gatgscora}
\end{figure*}

\begin{figure*}[!ht]
   
    \begin{subfigure}[b]{0.49\linewidth}
        \includegraphics[width=\linewidth]{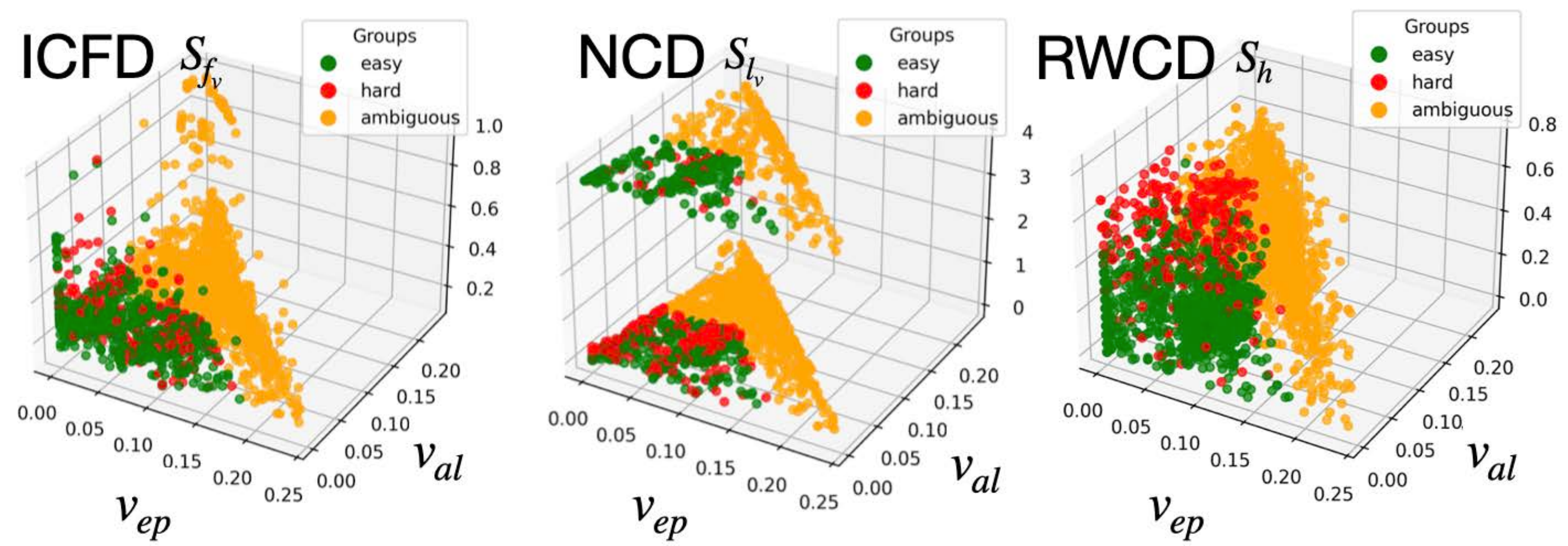}
    \caption{\gcn on \credit using built-in splits.}
\label{fig:correlation_scatter_partial_train_gcn_credit}
    \end{subfigure}
    \begin{subfigure}[b]{0.49\linewidth}
        \includegraphics[width=1.0\linewidth]{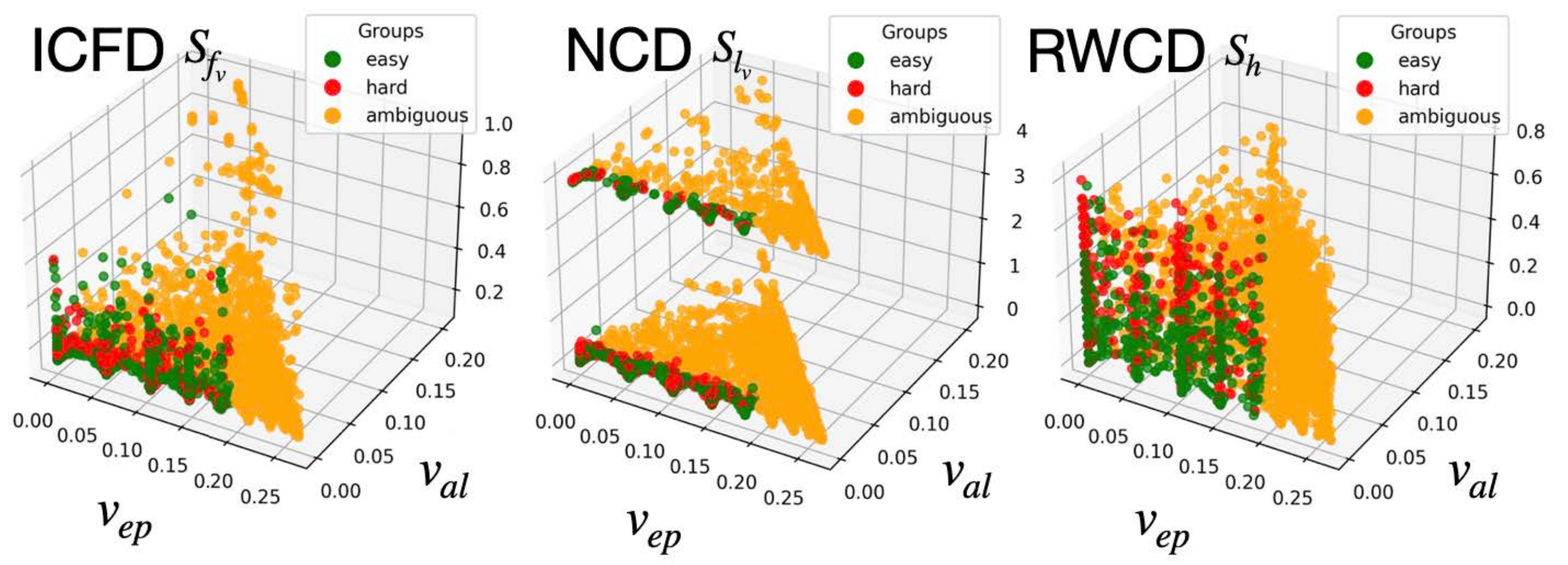}
    \caption{\gs on \credit using built-in splits.}
    \label{fig:correlation_scatter_partial_train_gs_credit}
    \end{subfigure}
   \caption{Diagnosing \gcn and \gs on \credit using built-in splits. }
\label{fig:gcngscredit} 
\end{figure*}

\begin{figure*}
   \begin{subfigure}[b]{0.49\linewidth}\includegraphics[width=1.0\linewidth]{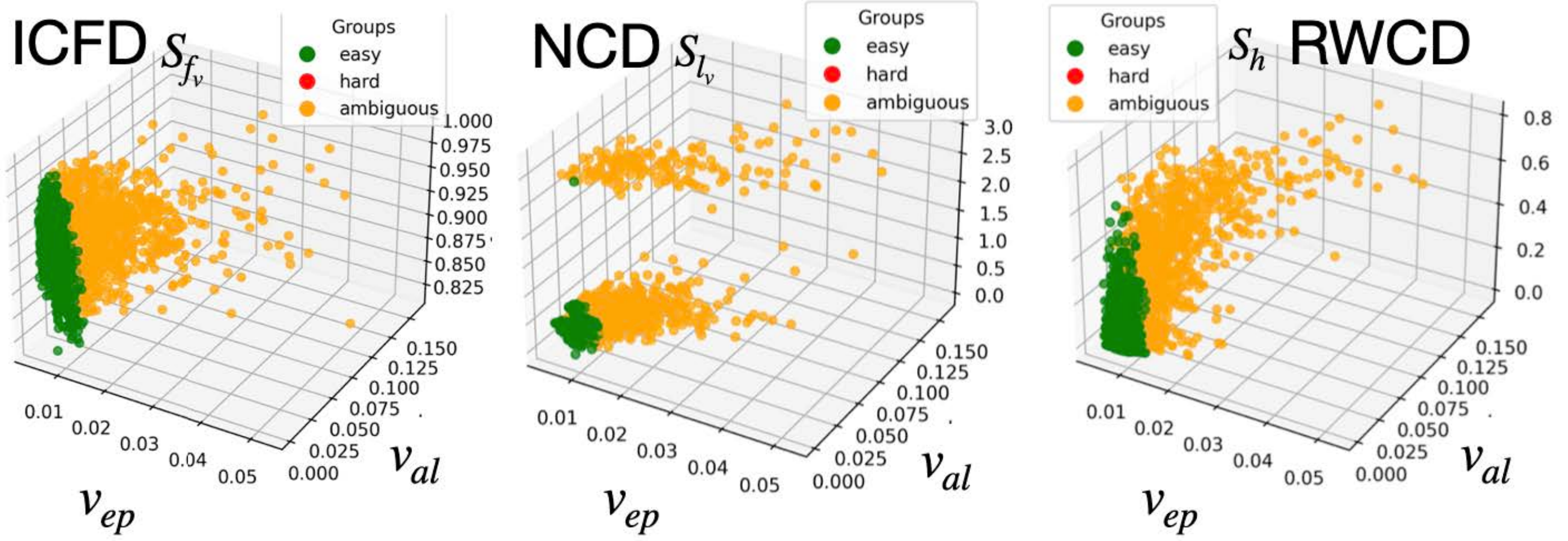}
    \caption{\gs on \cora trained with complete data.}
\label{fig:correlation_scatter_full_train_gs_cora}
\end{subfigure}
\begin{subfigure}[b]{0.49\linewidth}\includegraphics[width=1.0\linewidth]{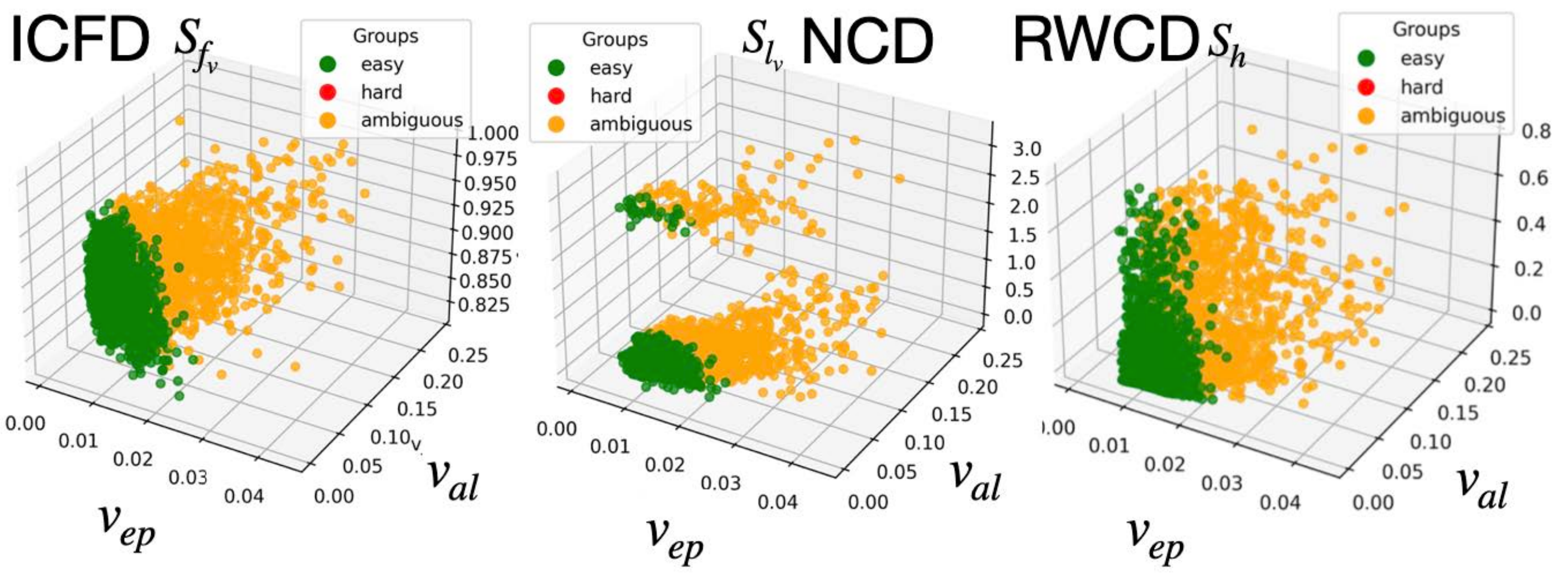}
    \caption{\mlp on \cora trained with complete data.}
   \label{fig:correlation_scatter_full_train_mlp_cora}
    \end{subfigure}
    \caption{Diagnosing \gs and \mlp on \cora in \ft setup.}
    \label{fig:gsmlpcoraft}
\end{figure*}

\begin{figure*}[!ht]

   \begin{subfigure}[b]{0.49\linewidth}\includegraphics[width=1.0\linewidth]{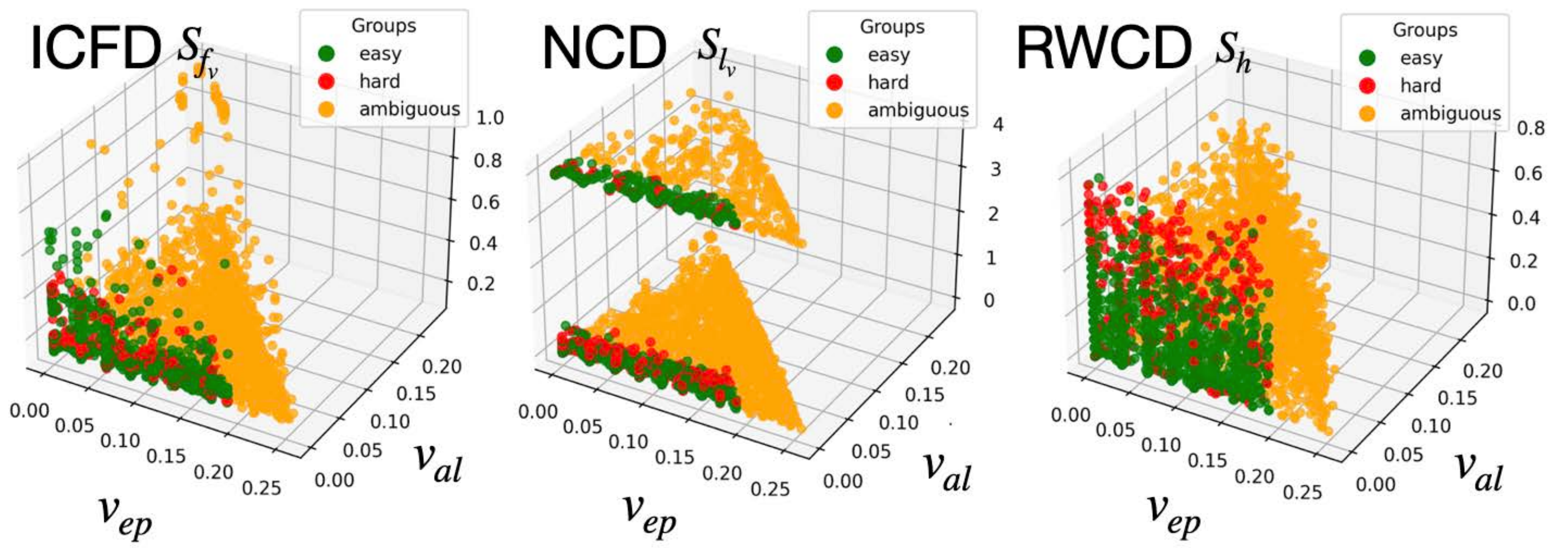}
   
    \caption{\gs trained on \credit with complete data.}
\label{fig:correlation_scatter_full_train_gs_credit}
\end{subfigure}
\begin{subfigure}[b]{0.49\linewidth}
\includegraphics[width=1.0\linewidth]{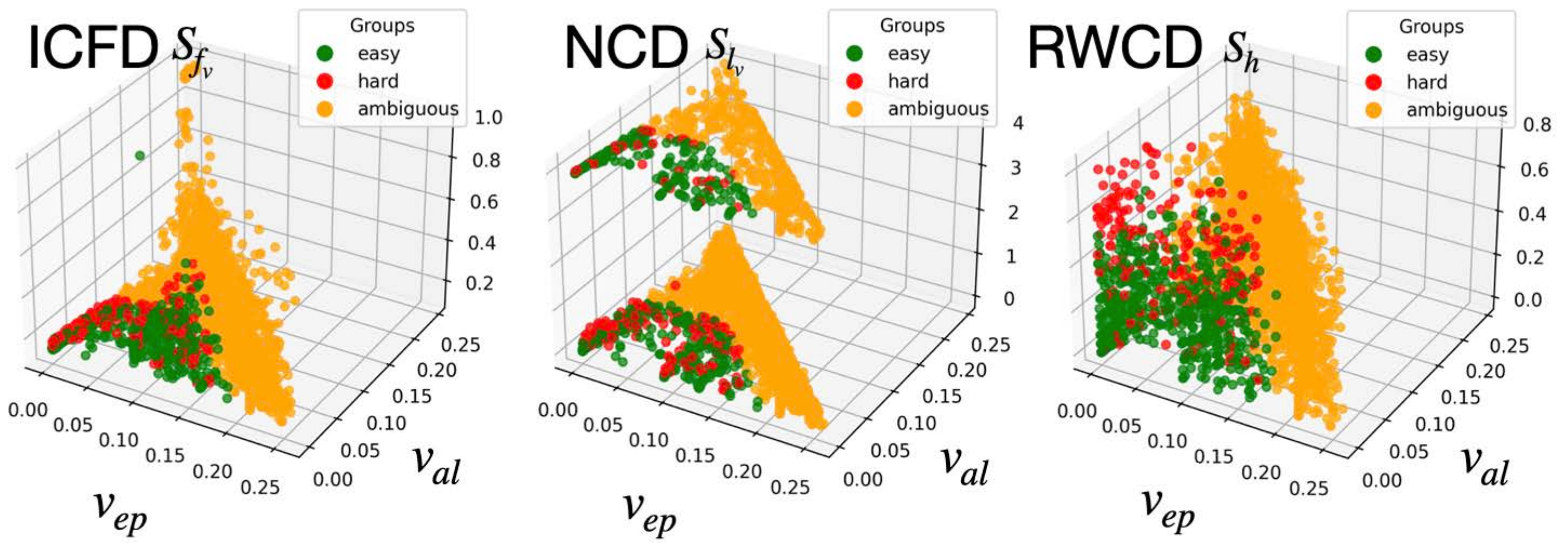}

    \caption{\mlp trained on \credit with complete data.}
\label{fig:correlation_scatter_full_train_mlp_credit}
\end{subfigure}
\caption{Diagnosing \gs and \mlp on \credit in \ft setup}
\label{fig:gsmlpcreditft}
\end{figure*}

\begin{figure*}[!ht]
    \begin{subfigure}[b]{0.49\linewidth}
   \includegraphics[width=1.0\linewidth]{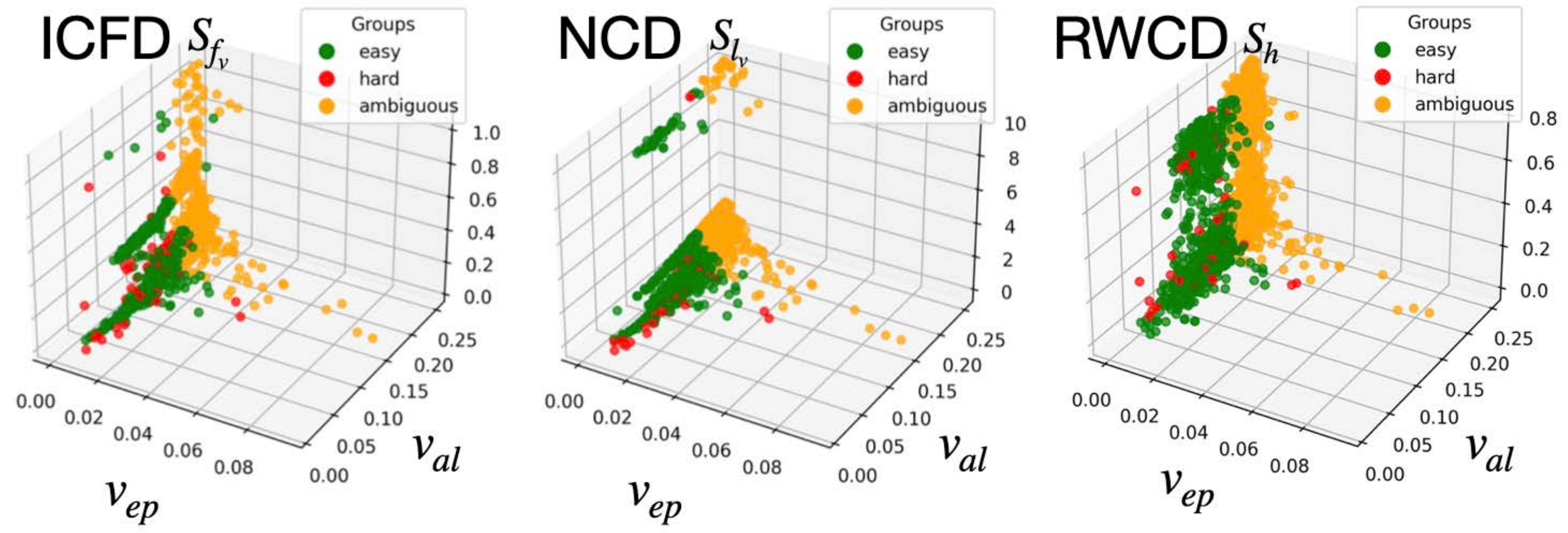}

    \caption{\gcn trained on \bitcoin with complete data.}
    \label{fig:correlation_scatter_full_train_gcn_bitcoin}
\end{subfigure}
 \begin{subfigure}[b]{0.49\linewidth}\includegraphics[width=1.0\linewidth]{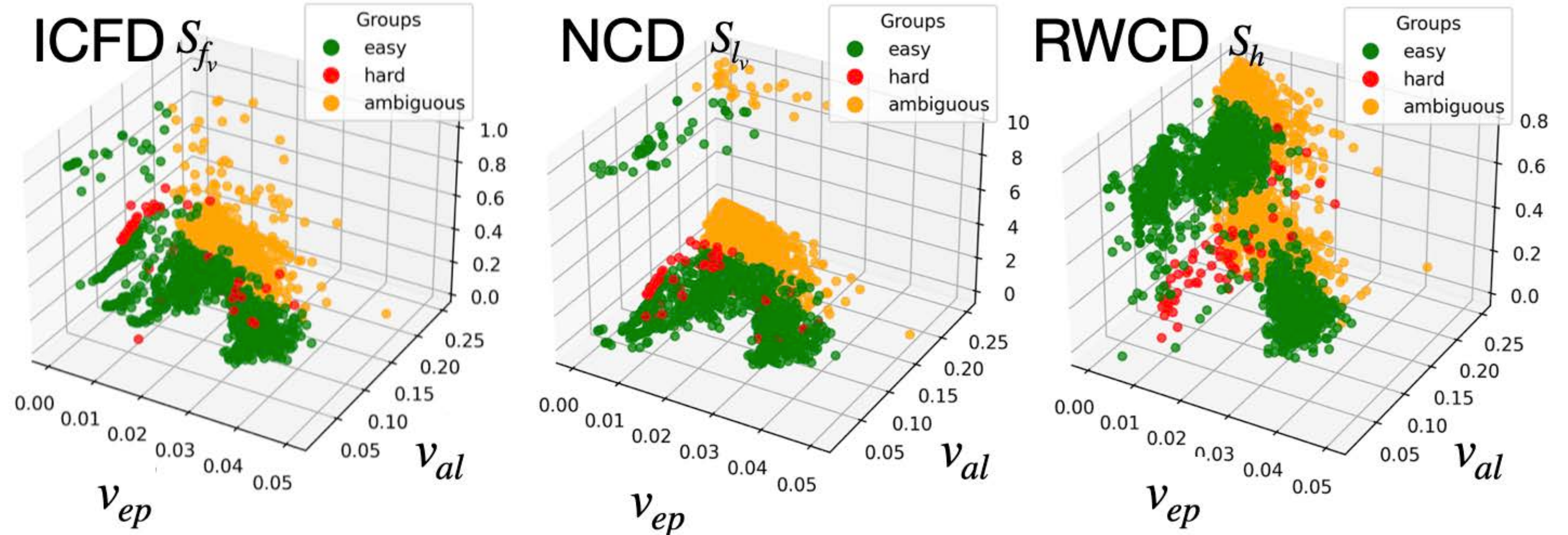}
   
    \caption{\gs trained on \bitcoin with complete data.}
\label{fig:correlation_scatter_full_train_gs_bitcoin}
\end{subfigure}
\caption{Diagnosing \gcn and \gs on \bitcoin in \ft setup}
\label{fig:gcngsbitcoinft}
\end{figure*}

We analyze the predictions of the models on \cora, \credit and \bitcoin datasets in correlation scatter plots (as in \cref{fig:gatgscora,fig:gsmlpcoraft,fig:gcngsbitcoinft,fig:gsmlpcreditft,fig:gcngscredit}). Each point represents a node in a given graph dataset, plotted with its epistemic uncertainty ($v_{ep}$ as $x$-axis), aleatoric uncertainty ($v_{al}$ as $y$-axis), with $z$-axis depicting one of the data profiling scores. Points are colored based on their assignments by \mymodel for all the nodes using the uncertainties defined in equation \ref{eq:def_vep}, \ref{eq:def_val} and \ref{eq:uncertainty_categorization}. 


% Note that, by definition, the same node of a given graph in the $3D$ correlation scatter plots share the same value along the $z$-axis, since the data-based profiling scores of a node are independent of the deployed models. In contrast, the coordinates of the same node along the epistemic uncertainty ($v_{ep}$, $x$-axis) and aleatoric uncertainty ($v_{al}$, $y$-axis) may vary depending on the training setup and the model used.
%\todo[inline]{For partial train did we not compute data profiling scores based on train set? If not, it is ok but let's not that point for ourselves to explore that later }

% \todo[inline]{The easy nodes find by one model is not for the other, high values on the profiling scores}
\paragraph{\pt set-up.} As shown in Table \ref{tab:nc_acc_pt_ft}, \gat and \gs achieve the same node classification accuracy on \cora in the \pt set-up. However, a comparison of the plots in Figures \ref{fig:correlation_scatter_partial_train_gat_cora} and \ref{fig:correlation_scatter_partial_train_gs_cora} reveals notable differences in their behavior. Specifically, \gs demonstrates higher prediction confidence, with \mymodel assigning $1296$ out of $2708$ nodes to the easy category and only $57$ to the hard category. In contrast, \gat exhibits lower confidence, with only $115$ nodes categorized as easy and $223$ as hard. This distinction is further reflected in lower values along the $v_{al}$ axis for \gs, indicating more stable prediction confidence. Notably, for both \gat and \gs, hard nodes show extremely low $v_{ep}$ values, suggesting that these nodes are predicted incorrectly and consistently so throughout training. 

Considering the data-centric scores on the $z$-axis, \gat particularly struggles with nodes that have high intra-class feature dissimilarity (\textsc{Icfd}) and neighborhood class divergence (\textsc{Ncd}) scores, indicating significant differences in input features and label distributions within their local neighborhoods, whereas \gs is able to partially learn  from or generalize to such nodes, despite their challenging characteristics.

In contrast for \credit (Figure \ref{fig:gcngscredit}) one can observe larger number of easy nodes that have high neighborhood class-divergence (\textsc{Ncd}) scores. Easy and hard nodes are better separated for \gcn than \gs where hardness seem to increase for \gcn with higher values of random walk class divergence (\textsc{Rwcd}) scores.
% Similar observations can be found by comparing the plots of \gcn and \gs on \credit in Figure \ref{fig:gcngscredit}, despite similar accuracy levels, \gcn classifies more nodes as easy and fewer as ambiguous and hard compared to \gs. With the same prediction confidence (as indicated by the value of $v_{al}$), \gcn also shows better consistency in its predicted reflected with lower value of $v_{ep}$. \gs shows over-confidence in the hard nodes indicated by the extremely low value of $v_{al}$ for the hard nodes. Interestingly, \gcn also predicts larger nodes correctly with
% higher confidence  (depicetd as easy nodes) with larger values for data-centric scores (scores on $z$-axis) as compared to \gs. 
%\todo[inline]{increase the font for $s_{f_v}$...}
%\todo[inline]{Did you use validation in partial train setting? I udnertsood that you just ran all modles till 100 epochs?}
\paragraph{\ft set-up.}As summarized in Table \ref{tab:nc_acc_pt_ft}, the \ft setup represents an upper-limit scenario in which all nodes in the graph are simultaneously used for training, validation, and testing.

In Figure~\ref{fig:gsmlpcoraft}, we compare \gs and \mlp, both achieving perfect accuracy on \cora in the \ft setup. For \gs , model-centric scores are compact and well-separated: easy nodes cluster tightly at the bottom-left, while ambiguous nodes form a broader but coherent region. In contrast, \mlp (Figure~\ref{fig:correlation_scatter_full_train_mlp_cora}) shows more dispersed clusters, with ambiguous nodes exhibiting higher variance in $v_{al}$ and several reaching $v_{ep}$ values up to $0.25$, suggesting overfitting to easy nodes and less stable predictions on ambiguous ones. With abundant training data, \mlp exhibits greater confidence in its correct predictions on nodes with high neighborhood-class divergence compared to \gs. One might then ask: \textit{is leveraging explicit graph structure always beneficial when sufficient training data is available?}

% \todo[inline]{The change of $1\%$ is incorrect}
% %\todo[inline]{the table appreance}
% \todo[inline]{use uncertainties 2d in the plot in the main}

The answer to the above question might not be trivial when we observe results on \textbf{\credit} dataset (Figure \ref{fig:gsmlpcreditft}).
Despite using the complete data for training none of the two models could not make correct predictions on all nodes. 
For both the best performing models (\gs and \mlp) there is no clear separation among easy and hard nodes on any of three axes. 
% \gs shows in Figure \ref{fig:correlation_scatter_full_train_gs_credit}a much tighter grouping: easy and hard nodes are concentrated below $v_{al} = 0.10$, while ambiguous nodes occupy the upper triangle with a clearer boundary. This indicates that \gs is more confident and consistent in its predictions. In contrast, \mlp shows in Figure \ref{fig:correlation_scatter_full_train_mlp_credit} spread on easy and hard nodes on both uncertainty axises, with $v_{ep}$ axis up to $0.25$ and $v_{al}$ around $0.10$, indicating inconsistent and low confidence. Ambiguous nodes dominate the region with $v_{ep} \in [0.00, 0.25]$ and $v_{al} \in [0.10, 0.25]$.

In contrast to \credit, we see a better separation of easy and hard nodes in \bitcoin (see Figure \ref{fig:gcngsbitcoinft}) dataset from \gs along the $z$-axis depicting random walk class divergence (\textsc{Rwcd}) score. This is not the case for the other best performing model i.e. \gcn.
When considering model-centric scores
\gcn shows compact and stable uncertainty estimates, with $v_{ep}$ mostly in the range $[0.00, 0.04]$ and $v_{al}$ between $[0.10, 0.25]$. Easy nodes form a dense cluster with low $v_{ep}$ and moderate $v_{al}$, hard nodes show only a slight increase in $v_{ep}$, and ambiguous nodes remain tightly grouped near-zero in $v_{ep}$ and close to the upper bound of $v_{al}$. Despite this concentration easy and hard nodes cannot be easily separated for \gcn making it harder to analyze the reasons for the observed hardness.
More detailed analysis of all remaining cases can be found in the Appendix.
%In contrast, \gs displays in Figure \ref{fig:correlation_scatter_full_train_gs_bitcoin} a wider spread, with $v_{ep}$ extending beyond $0.05$ and $v_{al}$ ranging from $0.00$ to $0.25$. Easy and hard nodes are more dispersed, and ambiguous nodes occupy a broader region along the $v_{ep}$ axis, reflecting greater inconsistency and less calibrated uncertainty estimates.

\subsection{Predicting Reliability in Predictions (RQ2)}
%\todo[inline]{What percentage of easy nodes did it predict correctly?}
%\todo[inline]{add the \% of correctly classified easy nodes}
We use \mymodel to predict the difficulty category \emph{easy}, \emph{ambiguous}, or \emph{hard} for previously unseen (test) nodes by applying a majority vote over their $K$-nearest neighbors in the model’s output representation space. This is done via a single forward pass over the subgraphs of the test nodes using the best-performing model checkpoint from the \pt setup. We fix $K=10$ for all experiments. Table~\ref{tab:prediction_acc} reports the categorization accuracy across three datasets—\cora, \credit, and \bitcoin—using four models: \gcn, \gat, \gs, and \mlp. \mymodel consistently demonstrates high accuracy in assigning difficulty labels to unseen nodes. Importantly, nodes predicted to be in the \emph{easy} category are typically associated with confident and stable predictions, suggesting that their model outputs can be trusted, even in the absence of ground truth.
%\todo[inline]{Disscuss!! Further on \credit?}

\begin{table}[!ht]
    \centering
    \begin{tabular}{lccc}
    \toprule 
    &\cora  &\credit &\bitcoin   \\\midrule
     \gcn  &$0.749$  &$0.679$ &$0.912$  \\
     \gat  &$0.793$  &$0.619$ &$0.846$  \\
     \gs  &$0.839$  &$0.630$ &$0.722$  \\
     \mlp  &$0.752$  &$0.587$ &$0.951$  \\
    \bottomrule
    \end{tabular}
    \caption{Categorization accuracy of test nodes achieved by \mymodel.}
    \label{tab:prediction_acc}
\end{table}

%We further analyze among all the test nodes, how much of the test nodes are categorized as easy, i.e., the prediction is categorized as reliable. 

\subsection{Identifying Semantic Errors (RQ3)}
We train \gs on our constructed graph \fbk, which includes both valid and intentionally corrupted (false) nodes (triples). As the model is exposed to predominantly correct data, we hypothesize that model-centric profiles from \mymodel can flag semantic inconsistencies, i.e., nodes whose learned patterns contradict established knowledge. Due to the scale of \fbk, we train with a batch size of $1024$. \mymodel classifies $34.9\%$ of test nodes as easy, $50\%$ as ambiguous, and $15.1\%$ as hard.

Crucially, \mymodel effectively isolates corrupted nodes: $97.3\%$ of the hard nodes correspond to flipped-label nodes, which encode contradictions to ground-truth facts. Overall, $48.9\%$ of all flipped-label nodes are categorized as hard, demonstrating \mymodel's ability to surface potential errors through model behavior profiling. 
% Given the large scale of the dataset \fbk, we trained \gs with batch size of $1024$ and report the categorization of the nodes in the Table \ref{tab:fbk_categorization_combined}. 

% In Table \ref{tab:fbk_categorization_combined}, we further analyze how well the corrupted nodes are identified by \mymodel. Among all the hard nodes, $97.3\%$ of the nodes are from the ones with which we flipped the labels. The nodes with flipped labels can be understood as our incorporated contradictions or deviation to the correct of typical knowledge embedded in the correctly labeled triples. We further calculate $48.9\%$ of the nodes with flipped labels are categorized as hard.

Further to qualitatively demonstrate the capability of \mymodel in identifying erroneous entries or deviations from typical knowledge in the constructed graph, we conducted a manual inspection of hard nodes flagged by the \mymodel. For instance, hard node \#272114 contains the corrupted triple $U = \langle$\textit{Paul Williams}, nominated for, \textit{Bad Boy Records}$\rangle$, which was originally $U = \langle$\textit{Paul Williams}, nominated for, \textit{A Star Is Born}$\rangle$. Similarly, another hard node \#201387 includes $U = \langle$\textit{Velvet Goldmine}, film/music, \textit{Carter Burwell}$\rangle$, which was incorrectly labeled as \textit{False} after label flipping. Both cases correspond to injected semantic errors and highlight the potential of \mymodel to surface such anomalies for further scrutiny.

To further investigate the behavior of \mymodel, we analyze the neighborhood of a specific hard node (index $201387$) in the \fbk graph. We randomly select six of its neighbors—two from each predicted difficulty category: \textit{easy}, \textit{ambiguous}, and \textit{hard}. Table~\ref{tab:fact_check_fbk} summarizes the semantic content of these neighbors, along with whether their labels were flipped during dataset construction.

As shown in Table~\ref{tab:fact_check_fbk}, all selected neighbors correspond to semantically valid triples, as indicated by their original (unflipped) labels. However, nodes with flipped labels are more frequently categorized as \textit{ambiguous} or \textit{hard}. These nodes are connected to the target node via shared entities like \textit{Velvet Goldmine} and \textit{Carter Burwell}, suggesting that \mymodel can identify semantic inconsistencies based on neighborhood patterns and learned difficulty profiles.

\begin{table}[!ht]
    \scriptsize

    \begin{tabular}{cccccc}
    \toprule
    Category & Entity1 & Relation & Entity2 & Label Flipped \\
    \midrule
    easy & Howl & Film/Music & Carter Burwell & No  \\
    easy & True Grit & Film/Music & Carter Burwell & No  \\
    ambiguous & bisexuality & Netflix genre & Velvet Goldmine & Yes  \\
    ambiguous & Eddie Izzard & Film & Velvet Goldmine & No  \\
    hard & film score & Music genre & Carter Burwell & Yes  \\
    hard & Velvet Goldmine & film location & London & Yes  \\
    \bottomrule
    \end{tabular}
    \caption{Example nodes from \fbk with predicted difficulty category, semantic content and whether an incorrect label was used during training.}
    \label{tab:fact_check_fbk}
\end{table}

\section{Conclusion}

In this work, we introduced \mymodel, a node profiling framework that captures fine-grained differences in graph model behavior by assigning interpretable scores to individual nodes. These scores combine data- and model-centric perspectives to identify failure patterns that aggregate metrics like accuracy or loss often miss. We demonstrated that the profiles generalize to unseen nodes, enabling prediction reliability assessment without ground-truth labels, and effectively detect injected errors in a knowledge graph. 
While our framework demonstrates strong effectiveness, a limitation lies in the current scope of data-centric profiles, which may not fully capture finer-grained structural and semantic variations across nodes. As future work, we plan to broaden these profiles and explore their application to tasks such as model selection, as well as extend the framework to dynamic and multi-label graph settings.

\bigskip
\bibliography{aaai2026}

%%%%%%%%%%%%%%%%%%%%%%%%%%%%%%%%%%%%%%%%%%%%%%%%%%%%%%%%%%%%%%%%%%%%%%%%%

% \input{../../ReproducibilityChecklist/LaTeX/ReproducibilityChecklist.tex}

\clearpage
\section{Appendix / supplemental material}

This appendix provides additional technical details, extended results, and supplementary discussions that support the analyses in the main paper. 

In Section \textit{Discussion on Multi-Label Datasets}, we extend our formulation of the neighborhood class divergence scorre to the multi-label node classification setting, complementing the single-label formulation presented in Section \textit{Our Proposed Framework} of the main paper. Section \textit{Model Hyperparameter Setting and Infrastructures} gives detailed descriptions of the datasets used and the hyperparameter settings for all models, supporting the experimental setup introduced in the main paper. In Section \textit{Additional Experimental Results}, we present further experimental results including node categorization performance across multiple random splits of the \bitcoin dataset, expanding the results presented for predicting reliability in predictions (RQ2). Section \textit{Correlation Analysis Between Data- and Model-based Node Profiling Scores} provides detailed analyses of model-centric and data-centric profiling scores for different models and datasets. We then include additional insights on model behavior during training in Section \textit{Training Behavior Analysis}, which compares the behavior differences of the models shown on different categories of the given datasets. Finally, in Section \textit{\fbk Without Flipping Labels}, we present an ablation study on the necessity of label flipping in the \fbk dataset.

\subsection{Discussion on Multi-Label Datasets}
\label{sec:multi_label_discussion}
As mentioned in the main paper, in the problem formulation, we focus on the multi-class node classification in this work, where one node can only belong to one class.  To include the possibility of multi-label node classification \cite{zhao2023multilabel}, where one node can belong to more than one class, the formulation of $S_{l_{v}}$ can be summarized as the average difference between $\mathcal{P}_{v}$ and averaged label distribution in the direct neighborhood of each class $y_v \in \mathcal{Y}_v$:

\begin{equation}
\label{eq:higher}
\begin{aligned}
S_{l_v} = \frac{1}{|\mathcal{Y}_v|} \sum_{y_v \in \mathcal{Y}_v} \sum_{c \in \mathcal{C}} \Big[ 
& \log(\mathcal{P}_v(c) + \epsilon) \\
& - \log(\mathcal{Q}_{y_v}(c)) 
\Big] \cdot (\mathcal{P}_v(c) + \epsilon)
\end{aligned}
\end{equation}

where we report the average \textsc{Ncd} to every class node $v$ belong to. Higher value of $S_{l_{v}}$ indicates a very different label distribution in the intermediate neighborhood of node $v$ compare to other nodes of the same class.

\subsubsection{Datasets}

\paragraph{\cora\cite{DBLP:journals/corr/YangCS16}} is a citation network in which each node represents a research article, and an edge connects two nodes if one article cites the other. Each node is labeled with the category of the corresponding article. Node features are binary word vectors ($0/1$), indicating whether a specific word appears in the abstract of the article.

\paragraph{\credit\cite{DBLP:journals/corr/abs-2102-13186}} contains customer-level financial data used to predict credit default risk. Attributes include demographics, credit history, and loan information, with some versions incorporating graph structures to represent user relationships. It supports tasks like credit scoring and risk assessment. We followed the pre-processing process in \cite{Olatunji_2023} to extract $3000$ nodes out of the original \credit dataset.

\paragraph{\bitcoin\cite{7837846}} is a signed, directed network comprising trading accounts, where each node corresponds to an individual account and each edge represents a trust rating issued by one user toward another. Edge weights range from $-10$ (indicating complete distrust) to $10$ (indicating complete trust). Following the procedure described in \cite{Olatunji_2023}, we construct node feature vectors based on received trust ratings and structural properties of the network, such as in-degree and out-degree.

\paragraph{\fbk} is a knowledge graph constructed from the original \fbkori dataset \cite{toutanova-chen-2015-observed}. Each node in the graph corresponds to a triple of the form $⟨u, e, v⟩$. Edges between nodes are established based on shared entities, i.e., an edge exists if two nodes share at least one common entity. Node labels indicate whether the corresponding node contains randomly replaced entity within the triple. nodes containing such errors are assigned to class $0$, while error-free nodes are assigned to class $1$. Furthermore, $30\%$ of the nodes have intentionally flipped labels, which introduces noise into the classification. We put the empirical evidence for the necessity of flipping the labels in the Appendix \fbk Without Flipping Labels. 

\subsection{Model Hyperparameter Setting and Infrastructures}
In this section, we summarize the hyperparameter configurations employed in this study for the collected machine learning models (shown in Table \ref{tab:hyperparams_2layer}) as well as for \mymodel. 

For all publicly-available models we retained the default settings provided in their respective GitHub repositories, because our objective is not to benchmark absolute performance but to probe the learning dynamics of each model on the target dataset. As noted in Section Training Behavior Analysis, we additionally sampled training $80$ checkpoints with uniform increment to minimize any effects from differences in training length.

\begin{table}[!ht]
\small
\setlength{\tabcolsep}{3pt}
\centering
\begin{tabular}{l l c c c}
\toprule
Dataset & Model & Hidden Dim & LR & Num Layers \\
\cora & GCN & 64 & 0.01 & 2 \\
\cora & GAT & 64 & 0.01 & 2 \\
\cora & MLP & 64 & 0.01 & 2 \\
\cora & GraphSAGE & 64 & 0.01 & 2 \\
\bitcoin & GCN & 256 & 0.001 & 2 \\
\bitcoin & GAT & 256 & 0.001 & 2 \\
\bitcoin & MLP & 256 & 0.001 & 2 \\
\bitcoin & GraphSAGE & 256 & 0.001 & 2 \\
\credit & GCN & 256 & 0.001 & 2 \\
\credit & GAT & 256 & 0.001 & 2 \\
\credit & MLP & 256 & 0.001 & 2 \\
\credit & GraphSAGE & 256 & 0.001 & 2 \\
\fbk & GraphSAGE & 256 & 0.001 & 2 \\
\bottomrule
\end{tabular}
\caption{Hyperparameter Settings. "LR" indicates learning rate.}
\label{tab:hyperparams_2layer}
\end{table}

For the choice of thresholds $c_{up}$ and $c_{low}$ in \mymodel, we used $0.75$ and $0.25$ for the datasets \cora, \credit, and \bitcoin. For \fbk, we used $c_{up}=0.6$, $c_{low}=0.4$.

All experiments were conducted on high-performance computing servers featuring 2 × Intel® Xeon® Gold 6140 CPUs @ 2.30GHz and 2 × AMD EPYC 7452 32-Core Processors, providing a combined infrastructure with a total of 72 Intel threads and 128 AMD threads.

\subsection{Additional Experimental Results}

We built three random splits on \bitcoin and report the performance of the four collected models in the \pt setup in the following Table \ref{tab:ex_std}:

\begin{table}[h!]
\setlength{\tabcolsep}{3pt}
\centering
\begin{tabular}{lcc}
\hline
 & Classification Accuracy & Profiling Accuracy \\
\hline
\gcn & $0.850\pm0.003$ &$0.878\pm0.080$ \\
\gat & $0.590\pm0.005$ &$0.942\pm0.068$\\
\gs &$0.835\pm0.020$ &$0.777\pm0.093$ \\
\mlp &$0.743\pm0.004$ &$0.965\pm0.011$ \\
\hline
\end{tabular}
\caption{Average node classification accuracy achieved by \gcn, \gat, \gs, and \mlp, and the node categorization accuracy achieved by \mymodel with standard deviation.}
\label{tab:ex_std}
\end{table}

Take the best performing model on \bitcoin in average \gcn as an example, we summarize the test categorization in Table \ref{tab:bitcoin_test_categorization}:

\begin{table}[h!]
\centering
\setlength{\tabcolsep}{4pt}
\begin{tabular}{lccc}
\hline
\textbf{Metric} & \textbf{split 0} & \textbf{split 1} & \textbf{split 2} \\
\hline
Total test nodes & $758$ & $758$ & $758$ \\
Easy &$138$ ($18.21\%$) & $176$ ($23.22\%$) & $109$ ($14.38\%$) \\
Ambiguous &$625$ ($81.13\%$) & $572$ ($75.46\%$) & $641$ ($84.56\%$) \\
Hard &$5$ ($0.66\%$) & $10$ ($1.32\%$) & $8$ ($1.06\%$) \\
\hline
\end{tabular}
\caption{Test node categorization in \bitcoin across three random splits using \mymodel.}
\label{tab:bitcoin_test_categorization}
\end{table}

\subsection{Correlation Analysis Between Data- and Model- based Node Profiling Scores}

We analyze the predictions on the three collected graph datasets in $3D$ correlation scatter plots in details in this section. The plots follow the same layout as in the main paper, where each point represents a node in a given graph dataset, plotted with its epistemic uncertainty ($v_{ep}$ as $x$-axis), aleatoric uncertainty ($v_{al}$ as $y$-axis), and one of the profiling scores ($S_{f_v}$, $S_{l_v}$, or $S_h$ as $z$-axis). Points are colored by group difficulty (easy in green, ambiguous in orange, or hard in red) as assigned by \mymodel using the uncertainties defined in equation \ref{eq:def_vep}, \ref{eq:def_val} and \ref{eq:uncertainty_categorization}.

% Note that, by definition, the same node of a given graph in the $3D$ correlation scatter plots share the same value along the $z$-axis, since the data-based profiling scores of a node are independent of the deployed models. In contrast, the coordinates of the same node along the epistemic uncertainty ($v_{ep}$, $x$-axis) and aleatoric uncertainty ($v_{al}$, $y$-axis) may vary depending on the training setup and the model used.

%This analysis reveals how uncertainty components contribute to model confidence and performance across different node characteristic subgroups.

\paragraph{\pt}

\begin{figure}
    \centering
    \includegraphics[width=1.0\linewidth]{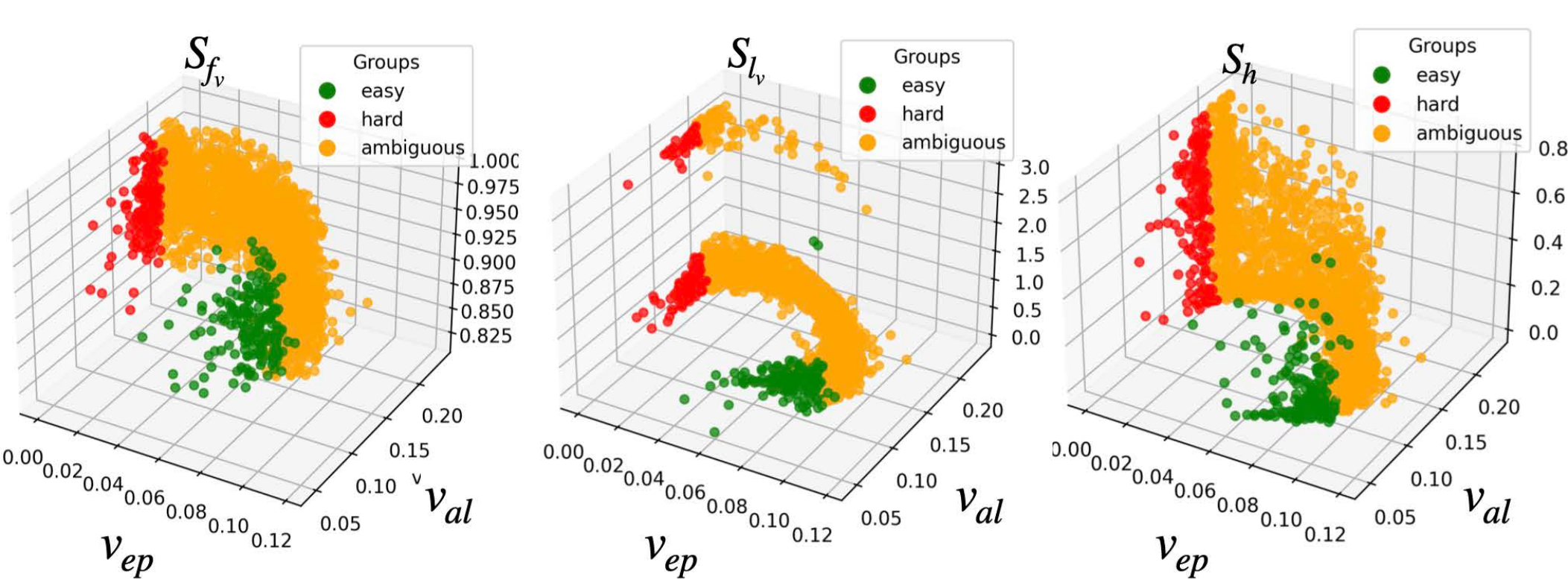}
    \caption{\pt correlation scatter plots of \gcn trained on \cora using built-in splits.}
    \label{fig:correlation_scatter_partial_train_gcn_cora}
\end{figure}

\begin{figure}
    \centering
    \includegraphics[width=1.0\linewidth]{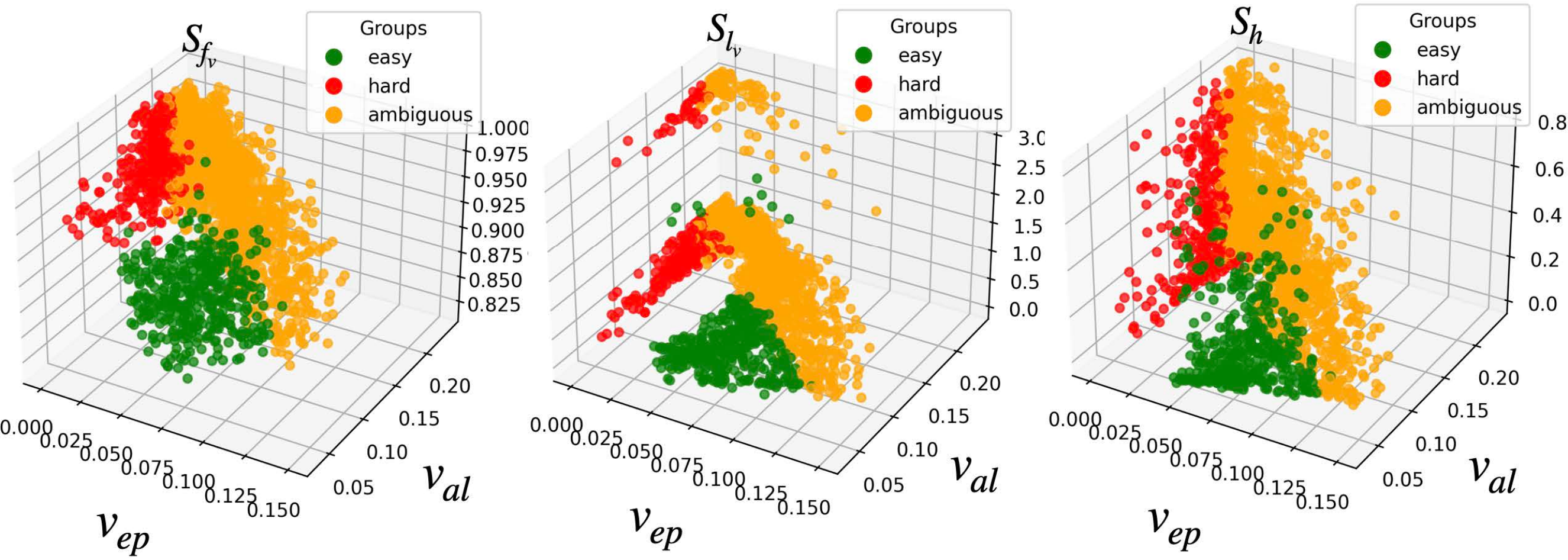}
    \caption{\pt correlation scatter plots of \mlp trained on \cora using built-in splits.}
    \label{fig:correlation_scatter_partial_train_mlp_cora}
\end{figure}

\paragraph{\cora} As shown in Figure \ref{fig:correlation_scatter_partial_train_gcn_cora}, \gcn exhibit a clear stratification of easy, ambiguous, and hard groups. Among all the nodes in the graph, easy samples identified by \mymodel are concentrated in regions of both low epistemic ($v_{ep}$) and aleatoric ($v_{al}$) uncertainty, indicating that \gcn not only predicts the correct labels but does so consistently with low variance. This suggests strongly confident and consistent predictions for these instances. Nodes identified as easy by \mymodel tend to have lower data-based node profiling scores compared to ambiguous and hard nodes. This suggests that the easy nodes are more similar  (in terms of features, local neighborhoods, and higher-order structures) to other nodes of the same class. Comparing with \mlp illustrated in Figure \ref{fig:correlation_scatter_partial_train_mlp_cora}, \mymodel also categorizes part of the nodes for \gcn with high \textsc{Icfd} and \textsc{Ncd} scores as ambiguous whereas \mlp solely depends on the input features and shows worse generalization for such challenging nodes.

% By contrast, as shown in Figure \ref{fig:correlation_scatter_partial_train_mlp_cora}, the \mlp model, lacking access to structural information, more nodes are categorized as hard. It also displays more scattered uncertainty distributions across all subgroups. Even easy samples show elevated aleatoric uncertainty (which means that the model is overall less confident for the correct predictions), highlighting the crucial role of graph structure in disentangling complex or ambiguous data instances and achieving confident predictions. The clearest clustering appears in the correlation scatter plot with $S_{f_v}$, which aligns with the fact that \mlp relies solely on node features to predict labels.

The gap in the distribution of $S_{l_v}$ across all collected models highlights a key characteristic of dataset \cora: the immediate neighborhood of the node in \cora tends to be either highly similar to the majority or distinctly different, with few cases where nodes partially share neighborhoods with others from the same class.

%#################################

\paragraph{\credit}

\begin{figure}[!ht]
    \centering
    \includegraphics[width=1.0\linewidth]{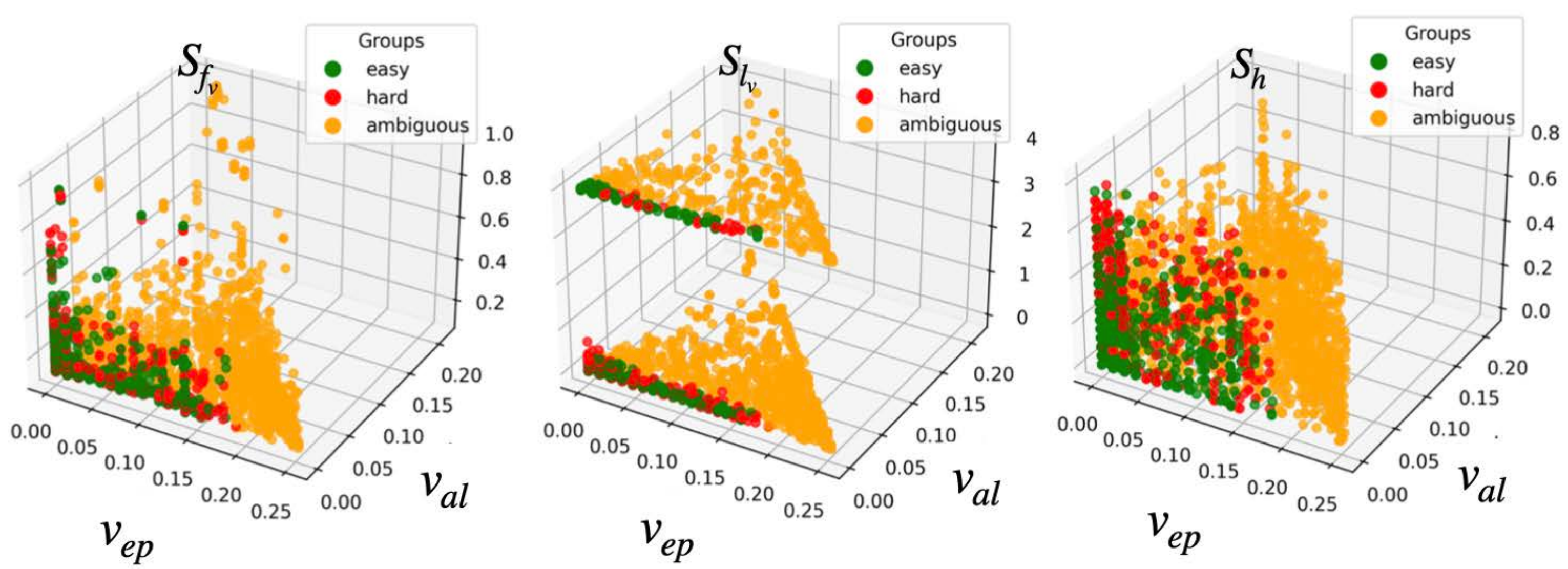}
    \caption{\pt correlation scatter plots of \gat trained on \credit using built-in splits.}
    \label{fig:correlation_scatter_partial_train_gat_credit}
\end{figure}

\begin{figure}
    \centering
    \includegraphics[width=1.0\linewidth]{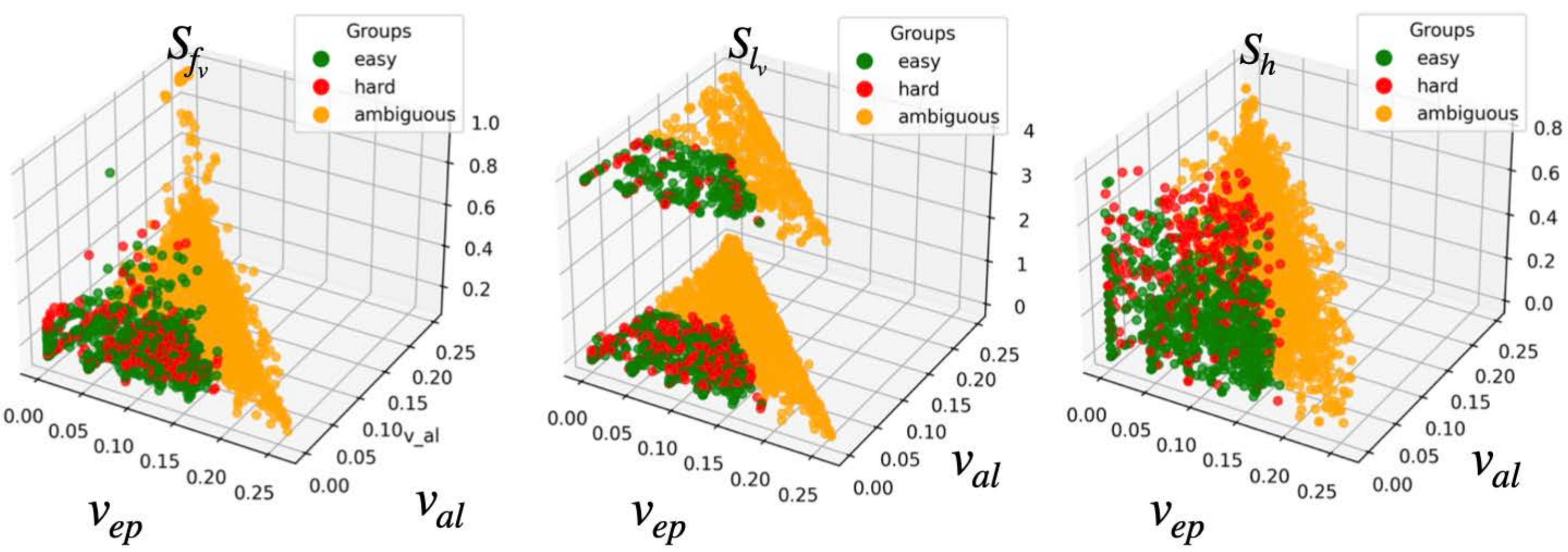}
    \caption{\pt correlation scatter plots of \mlp trained on \credit using built-in splits.}
    \label{fig:correlation_scatter_partial_train_mlp_credit}
\end{figure}

Figures \ref{fig:correlation_scatter_partial_train_gat_credit} and \ref{fig:correlation_scatter_partial_train_mlp_credit} present the correlation scatter plots for \gat and \mlp evaluated on the dataset \credit under the \pt setting. Interestingly, while the \mlp classifier achieves the highest test accuracy in this setup (as reported in Table \ref{tab:nc_acc_pt_ft}), it concurrently demonstrates the lowest consistency or confidence in its predictions on the unseen test nodes (as indicated by the categorization accuracy in Table \ref{tab:prediction_acc}). This suggests \mlp may overfit to the easy nodes in the train and test subsets and struggles to generalize to the ambiguous/hard nodes in the test subset. This observation further highlights that high test accuracy alone may not fully reflect the quality of the learned representations, and that evaluation frameworks like \mymodel, which reflects the confidence and consistency of the predictions are crucial for a more comprehensive assessment of model reliability.

% \todo[inline]{\mlp is performing best for \credit. Any thoughts?
% Do we see also differences in behavior of differenet models even if they have similar performance? }
% \mlp shows poorer separation between the groups and assign more nodes as ambiguous. This suggests that \mlp does not produce well-calibrated uncertainty estimates and often exhibits low variance for samples later labeled as ambiguous.

%#################################

\paragraph{\bitcoin}

\begin{figure}[!ht]
    \centering
    \includegraphics[width=1.0\linewidth]{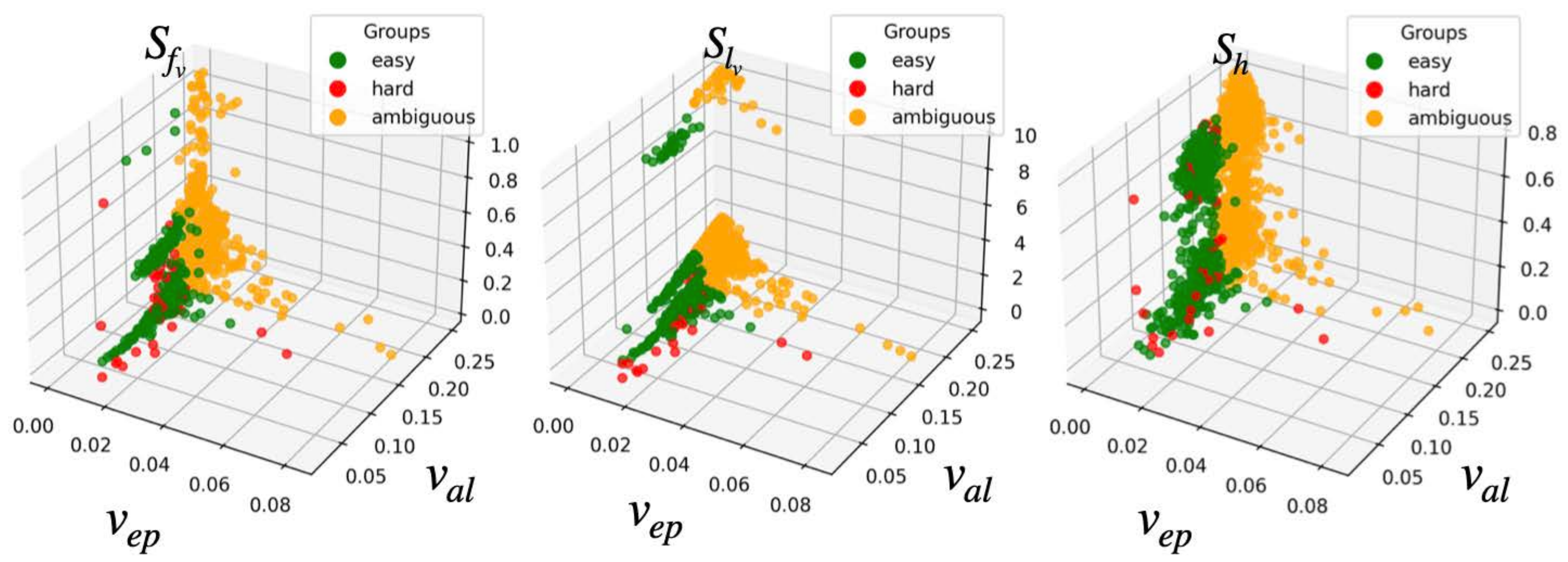}
    \caption{\pt correlation scatter plots of \gcn trained on \bitcoin.}
    \label{fig:correlation_scatter_partial_train_gcn_bitcoin}
\end{figure}

\begin{figure}[!ht]
    \centering
    \includegraphics[width=1.0\linewidth]{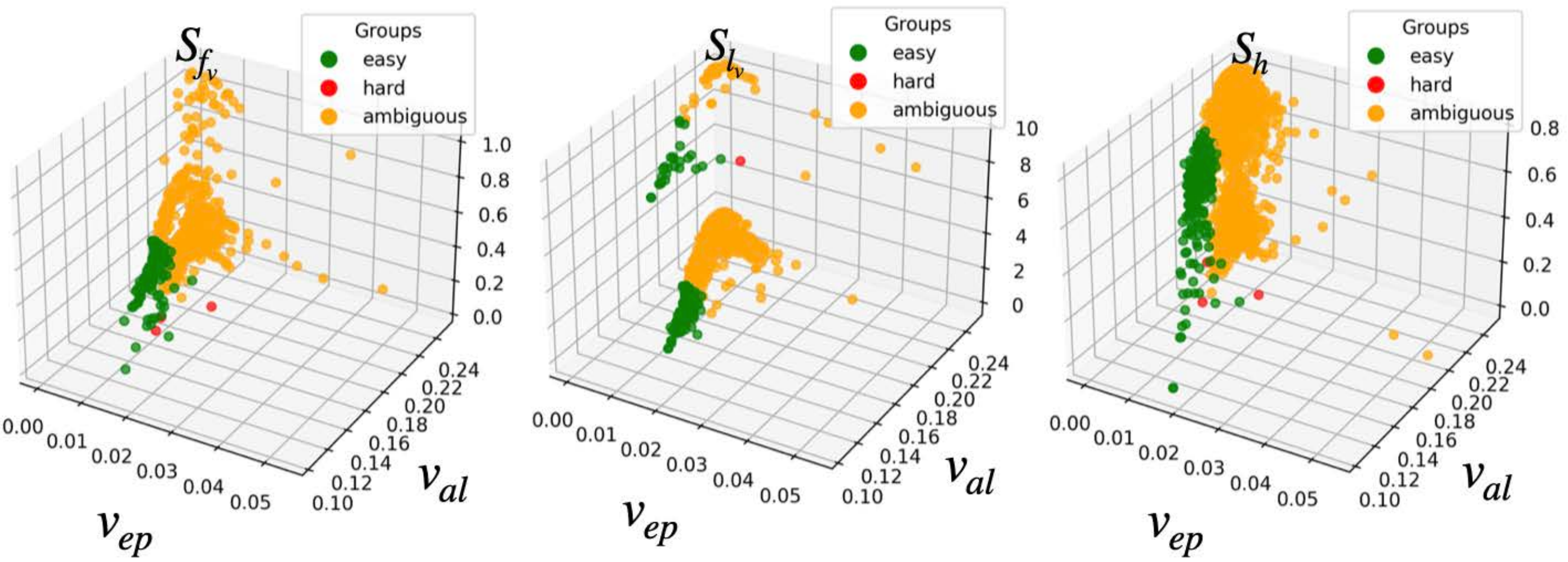}
    \caption{\pt correlation scatter plots of \gat trained on \bitcoin.}
    \label{fig:correlation_scatter_partial_train_gat_bitcoin}
\end{figure}

\begin{figure}[!ht]
    \centering
    \includegraphics[width=1.0\linewidth]{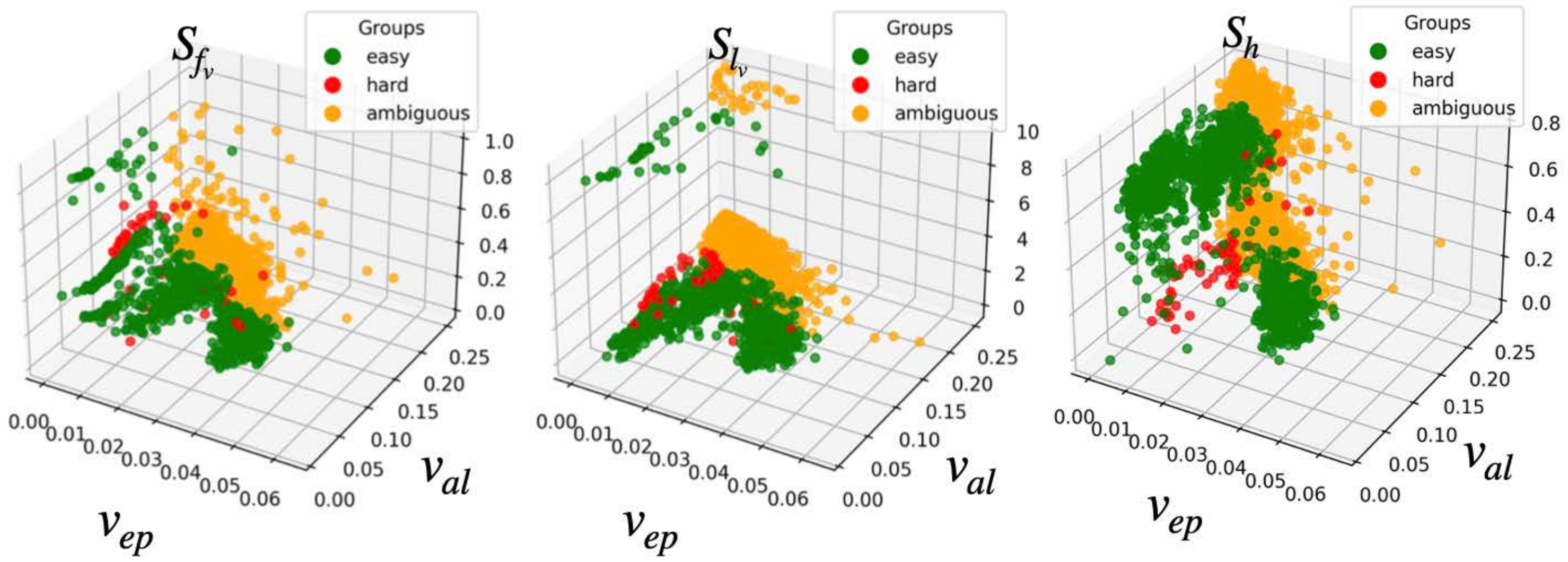}
    \caption{\pt correlation scatter plots of \gs trained on \bitcoin.}
    \label{fig:correlation_scatter_partial_train_gs_bitcoin}
\end{figure}

\begin{figure}[!ht]
    \centering
    \includegraphics[width=1.0\linewidth]{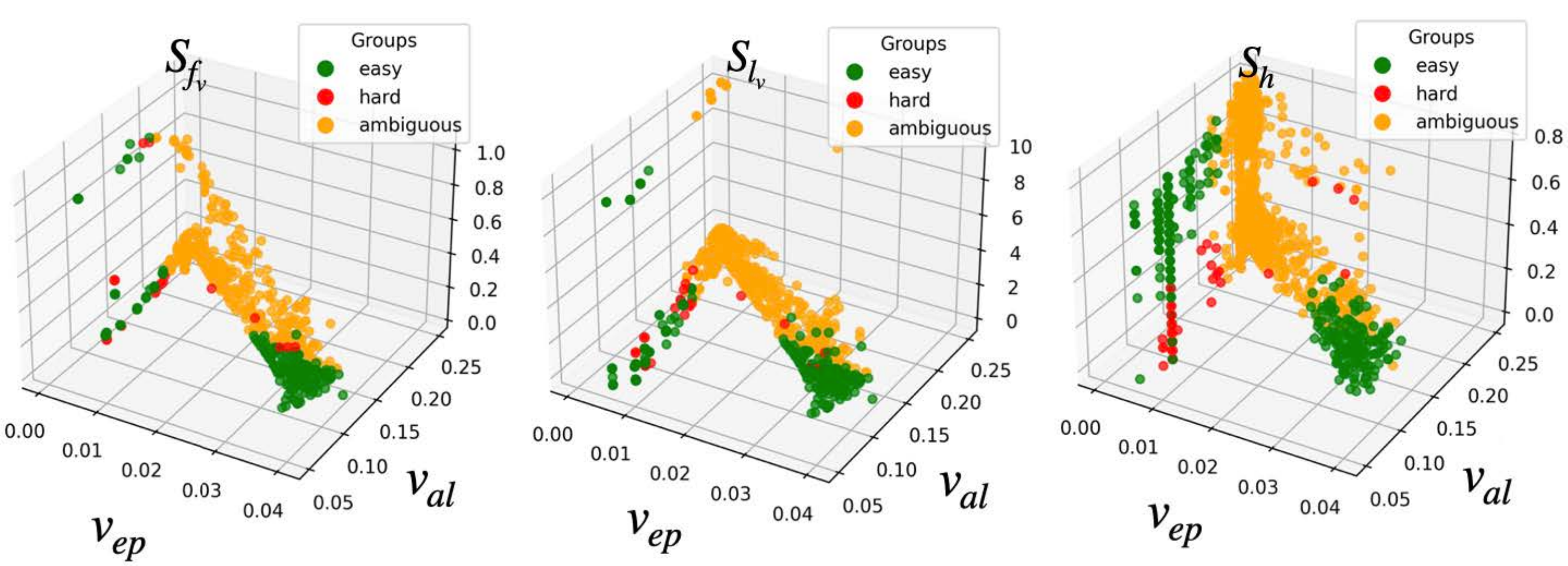}
    \caption{\pt correlation scatter plots of \mlp trained on \bitcoin.}
    \label{fig:correlation_scatter_partial_train_mlp_bitcoin}
\end{figure}

Figures \ref{fig:correlation_scatter_partial_train_gcn_bitcoin} to \ref{fig:correlation_scatter_partial_train_mlp_bitcoin} present the correlation scatter plots for the collected models on the \bitcoin dataset. Compared to the other two datasets, nodes categorized as easy, ambiguous, and hard by \mymodel tend to exhibit higher $v_{al}$ values and lower $v_{ep}$ values when using \gcn, \gat, and \mlp. This pattern suggests that model confidence on \bitcoin is generally lower than on other datasets. Even when predictions are correct, the predicted probability for the true label is not substantially higher than for the alternative class. 

In contrast, the scatter plots for \gs exhibit a noticeably different distribution. When examined alongside data-centric node profiling, we observe that many nodes categorized as easy by \mymodel tend to have low $\textsc{Icfd}$ and \textsc{Ncd} values. These characteristics suggest that such nodes are structurally and semantically more aligned with the rest of the nodes in their class. In comparison, the same nodes are often categorized as ambiguous when evaluated with other models. Additionally, \mymodel assigns only a small fraction of nodes to the hard group for \gs, reflecting a more consistent and confident prediction pattern. These observations indicate that the node subsets identified as easy by \mymodel exhibit properties that \gs can generalize to more effectively, underscoring its adaptability to the structural nuances of the \bitcoin dataset.

%%%%%%%%%%%%%%%%%%%%%%%%%% Full train scatters %%%%%%%%%%%%%%%%%%%%%%%%%%
\paragraph{\ft}

\begin{figure}[!ht]
    \centering
    \includegraphics[width=1.0\linewidth]{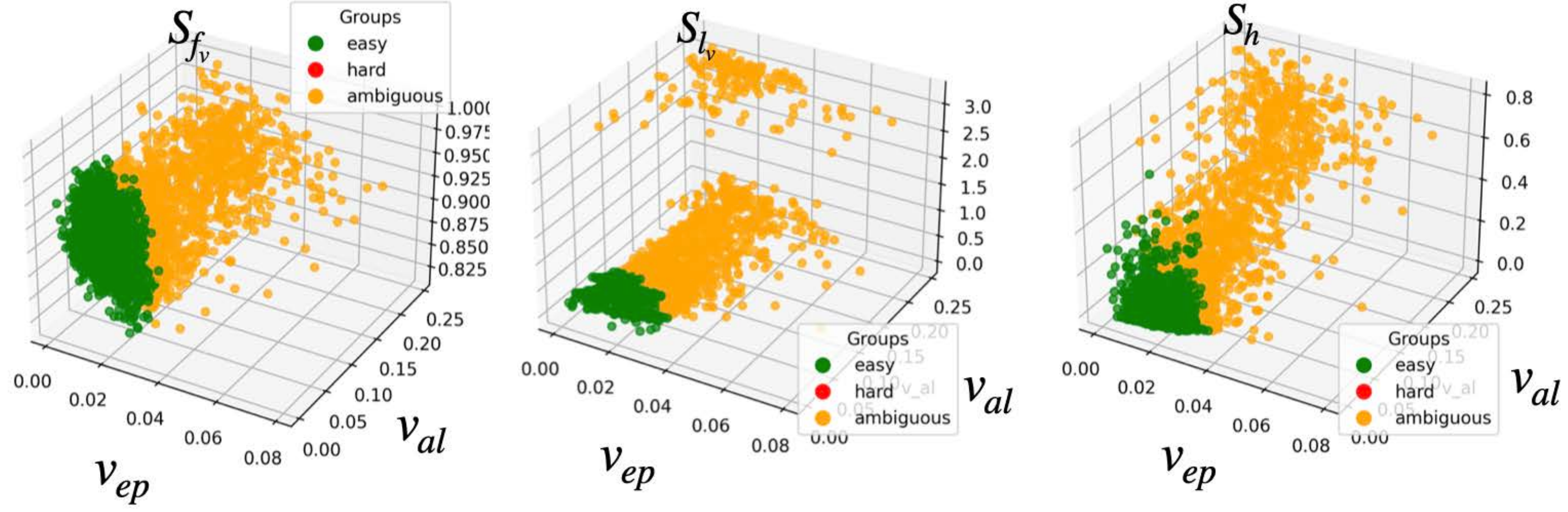}
    \caption{\ft correlation scatter plots of \gcn trained on \cora.}
    \label{fig:correlation_scatter_full_train_gcn_cora}
\end{figure}

\begin{figure}[!ht]
    \centering
    \includegraphics[width=1.0\linewidth]{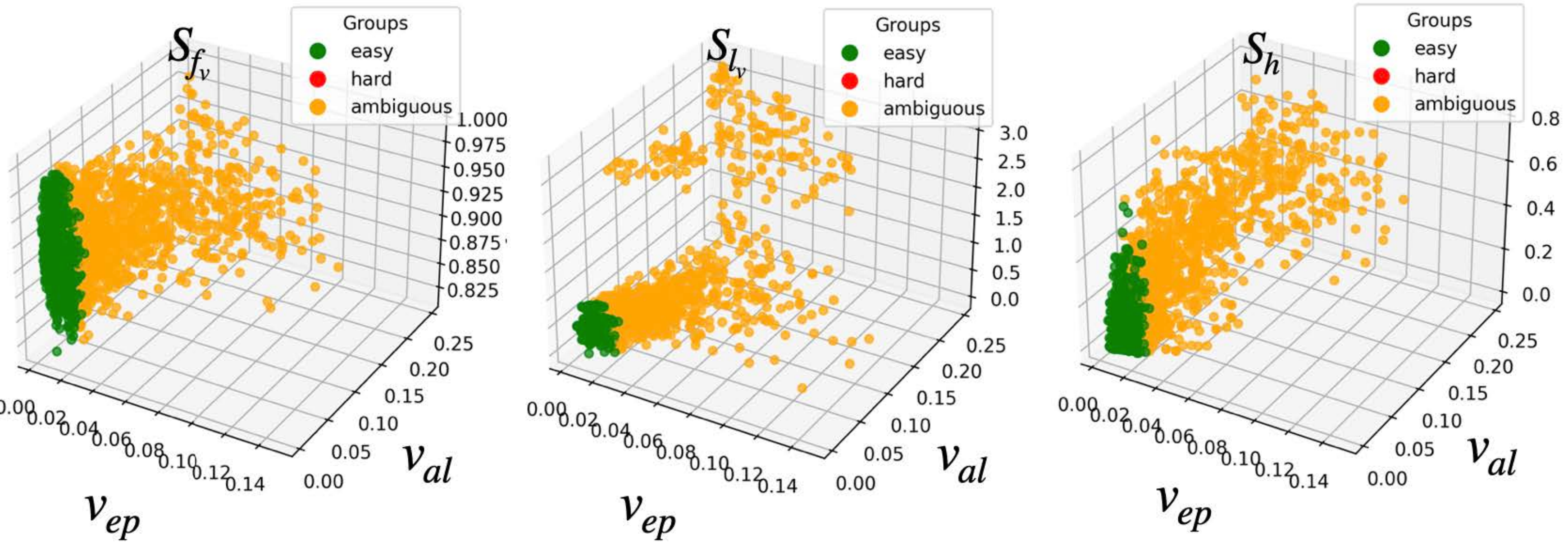}
    \caption{\ft correlation scatter plots of \gat trained on \cora.}
    \label{fig:correlation_scatter_full_train_gat_cora}
\end{figure}

\paragraph{\cora} As shown in Figure~\ref{fig:correlation_scatter_full_train_gcn_cora}, the \gcn model trained under the \ft setting exhibits improved performance compared to the \pt setting (Figure~\ref{fig:correlation_scatter_partial_train_gcn_cora}). Specifically, a larger number of nodes with high $\textsc{Icfd}$ and $\textsc{Rwcd}$ scores are categorized as easy by \mymodel. Additionally, nodes categorized as easy in \pt with low $\textsc{Ncd}$ are predicted with lower variance in \ft, as reflected in reduced $v_{ep}$ values, demonstrating substantially lower variance in prediction. However, the ambiguous nodes (orange) appear more scattered than the ambiguous nodes in the \pt setup, though they remain relatively distinct from the easy group. This dispersion suggests that under \ft, \gcn may overfit to the easy instances, while still struggling to generalize to more uncertain, ambiguous cases. As a result, \gcn may rely heavily on memorized patterns from confidently learned nodes, leading to reduced robustness when encountering structurally or semantically complex inputs.

As shown in Figure \ref{fig:correlation_scatter_full_train_gat_cora}, \gat exhibits a similar trend to \gcn in terms of data-centric node profiling. However, the ambiguous nodes (orange) in \gat are more widely spread along the $v_{ep}$ axis compared to \gcn. This suggests that although \gat achieves low uncertainty for easy nodes (green), its attention mechanism may inadvertently introduce noise or overemphasize less informative neighbors, resulting in less stable representations for ambiguous cases. Nevertheless, the distinction between easy and ambiguous node groups remains largely preserved.

\begin{figure}[!ht]
    \centering
    \includegraphics[width=1.0\linewidth]{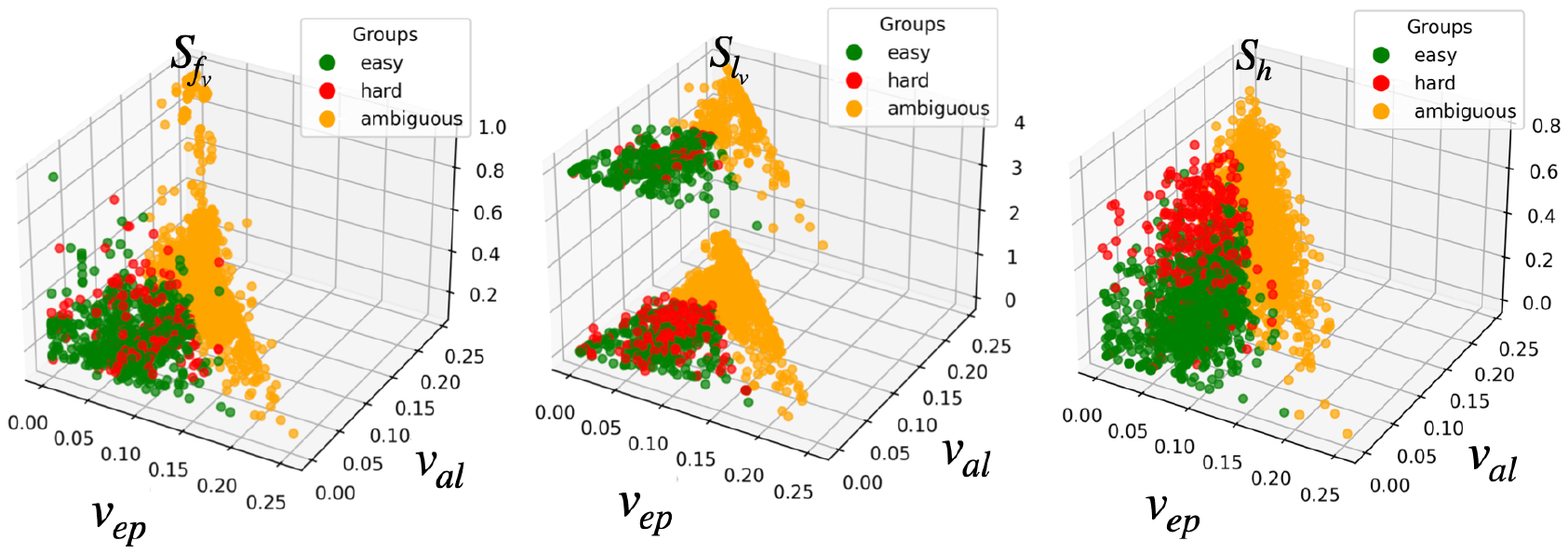}
    \caption{\ft correlation scatter plots of \gcn trained on \credit.}
    \label{fig:correlation_scatter_full_train_gcn_credit}
\end{figure}

\begin{figure}[!ht]
    \centering
    \includegraphics[width=1.0\linewidth]{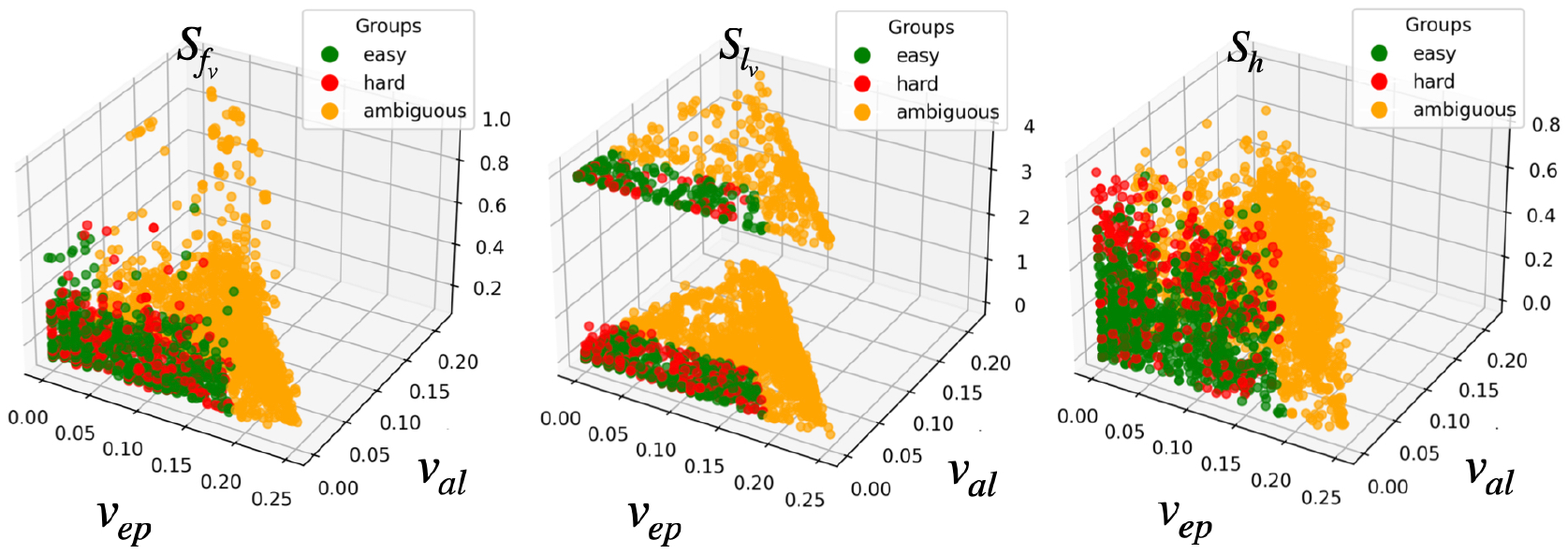}
    \caption{\ft correlation scatter plots of \gat trained on \credit.}
    \label{fig:correlation_scatter_full_train_gat_credit}
\end{figure}

\paragraph{\credit} Figures~\ref{fig:correlation_scatter_full_train_gcn_credit} and~\ref{fig:correlation_scatter_full_train_gat_credit} present the results of the correlation analysis conducted on the \credit dataset using \gcn and \gat under the \ft setup. Compared to the \pt setting, \mymodel categorizes more nodes with high $\textsc{Icfd}$, $\textsc{Ncd}$, and $\textsc{Rwcd}$ scores as easy for both \gcn and \gat, with comparably similar ranges of uncertainties. This observation suggests that, when provided with sufficient training data, \gcn and \gat generalize better to the challenging nodes with stable predictions.

\paragraph{\bitcoin}

\begin{figure}[!ht]
    \centering
    \includegraphics[width=1.0\linewidth]{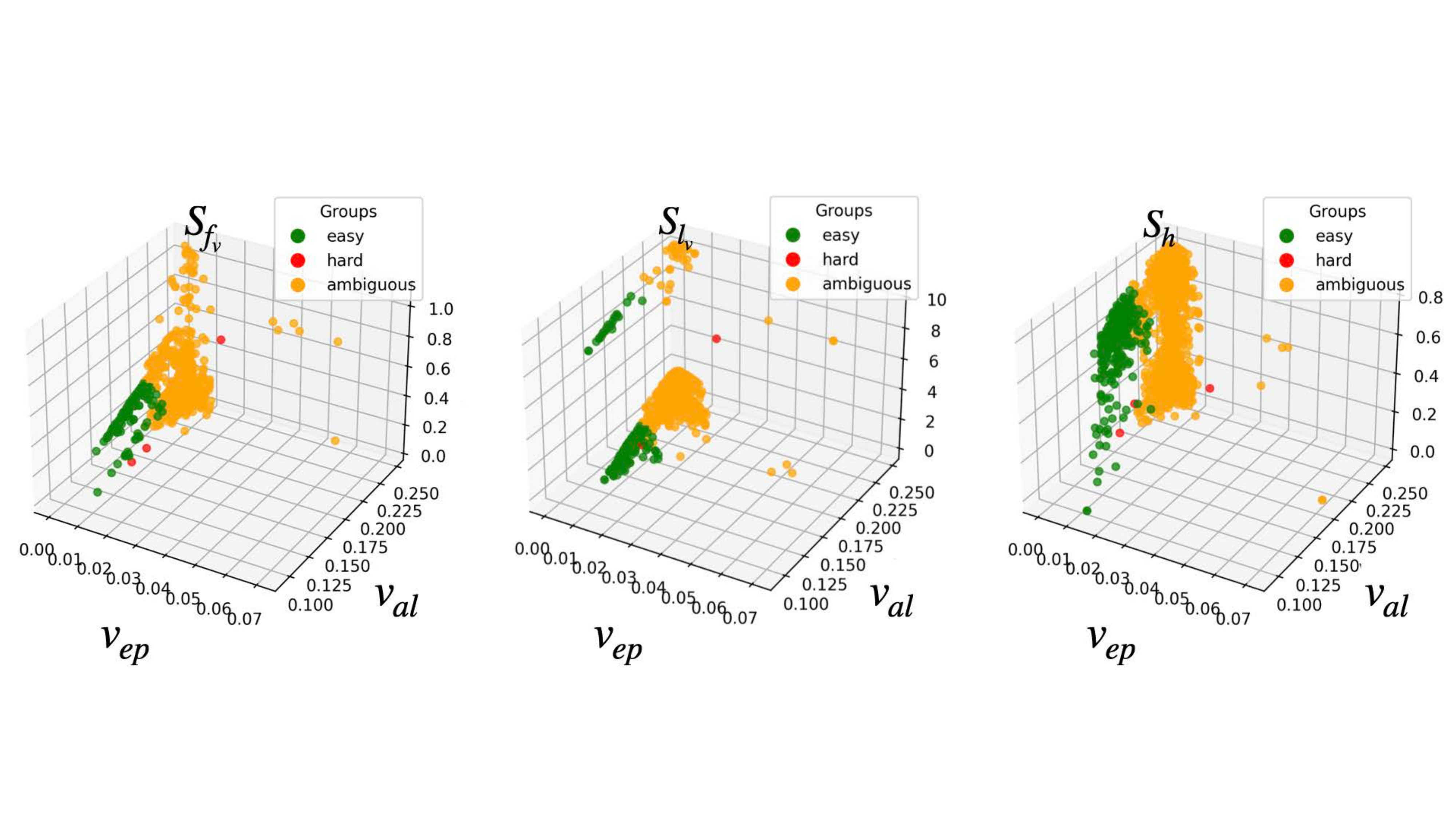}
    \caption{\ft correlation scatter plots of \gat trained on \bitcoin.}
    \label{fig:correlation_scatter_full_train_gat_bitcoin}
\end{figure}

\begin{figure}[!ht]
    \centering
    \includegraphics[width=1.0\linewidth]{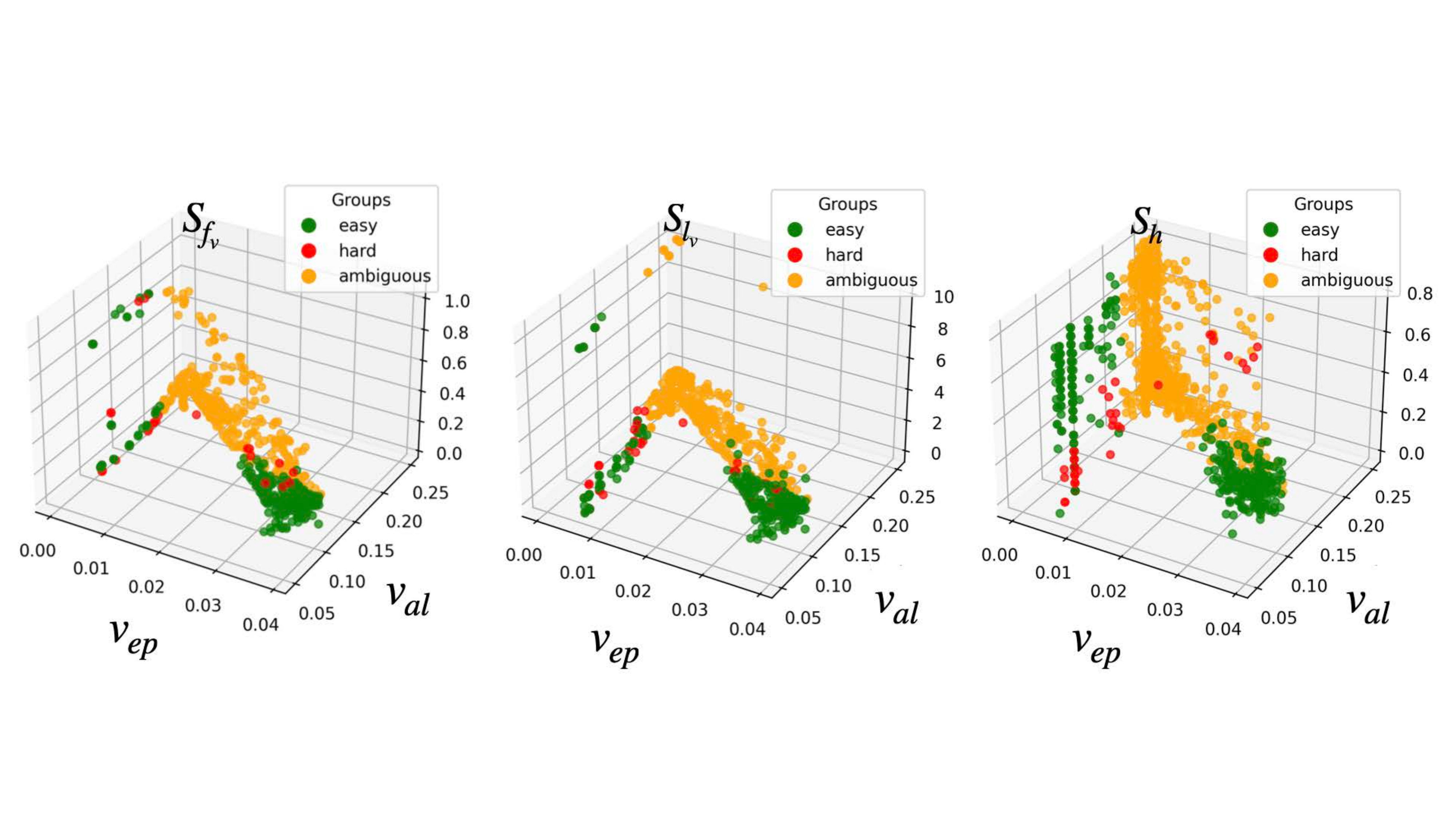}
    \caption{\ft correlation scatter plots of \mlp trained on \bitcoin.}
    \label{fig:correlation_scatter_full_train_mlp_bitcoin}
\end{figure}

Figures~\ref{fig:correlation_scatter_full_train_gat_bitcoin} and~\ref{fig:correlation_scatter_full_train_mlp_bitcoin} show the 3D correlation scatter plots for the \bitcoin dataset under the \ft setting using \gat and \mlp. According to \mymodel's categorization, more nodes are labeled as easy and fewer as hard for \gat, and the overall spread of the ambiguous group becomes tighter along the uncertainty axis compared to \gat in \pt. This indicates that \gat leverages the additional supervision to refine its predictions.

In contrast, \mlp, while having access to structural statistics encoded in the input features, such as degree and edge weight distributions, does not explicitly model the relational graph structure. As a result, it exhibits only subtle changes in both uncertainty measures under the \ft setting. There is a slight reduction in $v_{ep}$ and a modest tightening of the ambiguous cluster in the $\textsc{Icfd}$ and $\textsc{Rwcd}$ plots. However, the separation between nodes categorized as easy and hard remains weak.

%%%%%%%%%%%%%%%%%%%%%%%%%%%%%%%%%%%%%%%%%%%%%%%%%%%%%%%%%%%%%%%%%%%%%%%%%%%%%%%%%%%%%%%%%%%%%%%%%%%%%%%%

\subsection{Training Behavior Analysis}
\label{sec:TrainingBehavior}

In this section, we analyze the behavior of the collected models in the \pt setup with respect to nodes categorized as easy, ambiguous, and hard—represented in green, orange, and red, respectively. Specifically, we track and summarize the average training loss for each node category across sampled checkpoints, visualizing the results using line plots. In all plots, the $x$-axis denotes the indices of the sampled training checkpoints, while the $y$-axis represents the corresponding average training loss. These subgroups were determined based on the ground-truth categorization of all nodes defined in equation \ref{eq:def_vep}, \ref{eq:def_val} and \ref{eq:uncertainty_categorization}, and the probabilities were computed via forward passes on the graph datasets without backpropagation.

\label{AppendixSection:training_dym_analysis}

%%%%%%%%%%%%%%%%%%%%%%%%%%%%%%%%%%%%%%%%%%%%%%%%%%%%%%%%%%%%%%%%%%%%%%%%
\paragraph{\cora}

\begin{figure}[!ht]
    \centering
    \begin{subfigure}[b]{0.48\linewidth}
        \includegraphics[width=\linewidth]{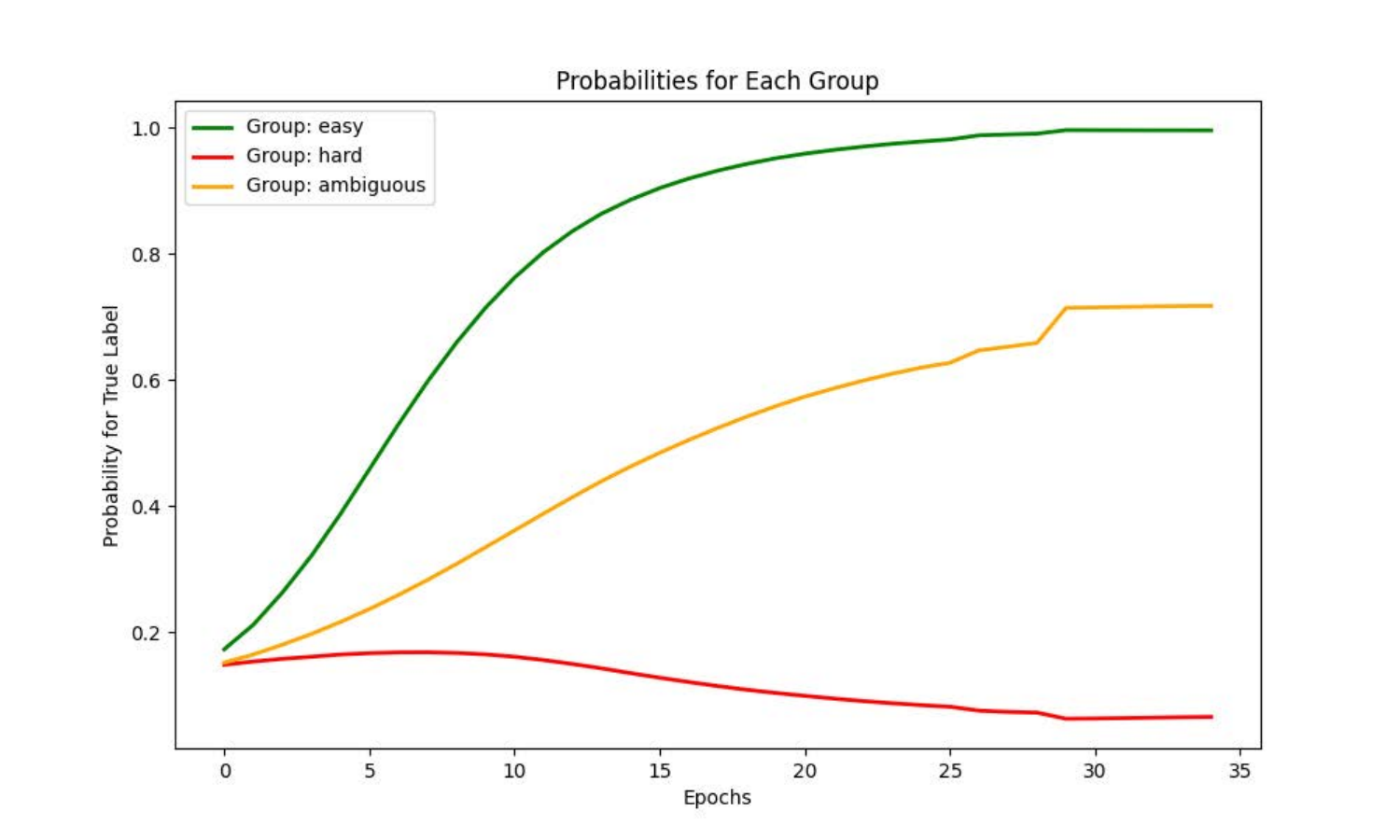}
        \caption{cora gcn}
    \end{subfigure}
    \hspace{0.01\linewidth}
    \begin{subfigure}[b]{0.48\linewidth}
        \includegraphics[width=\linewidth]{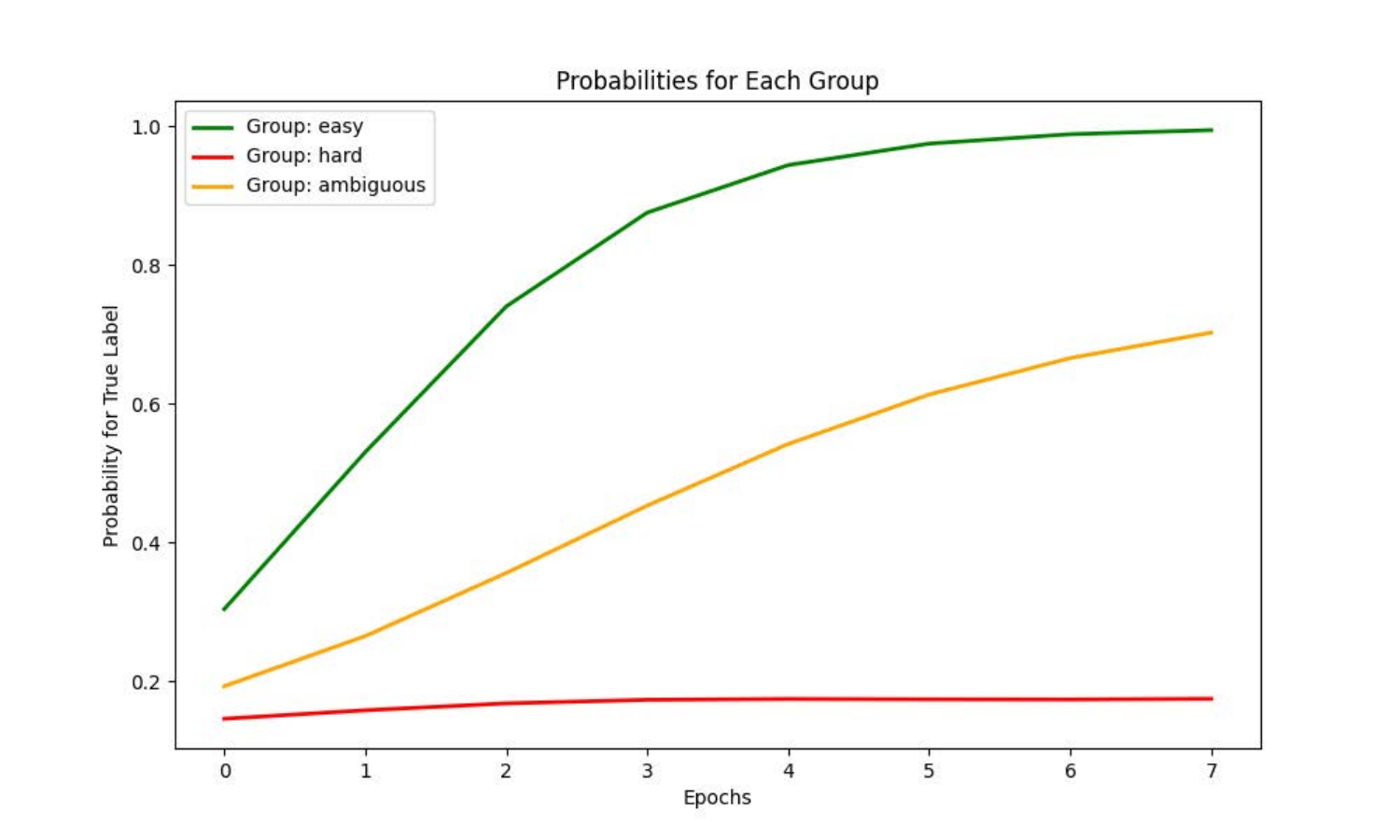}
        \caption{cora gat}
    \end{subfigure}
    \vspace{2mm}
    \begin{subfigure}[b]{0.48\linewidth}
        \includegraphics[width=\linewidth]{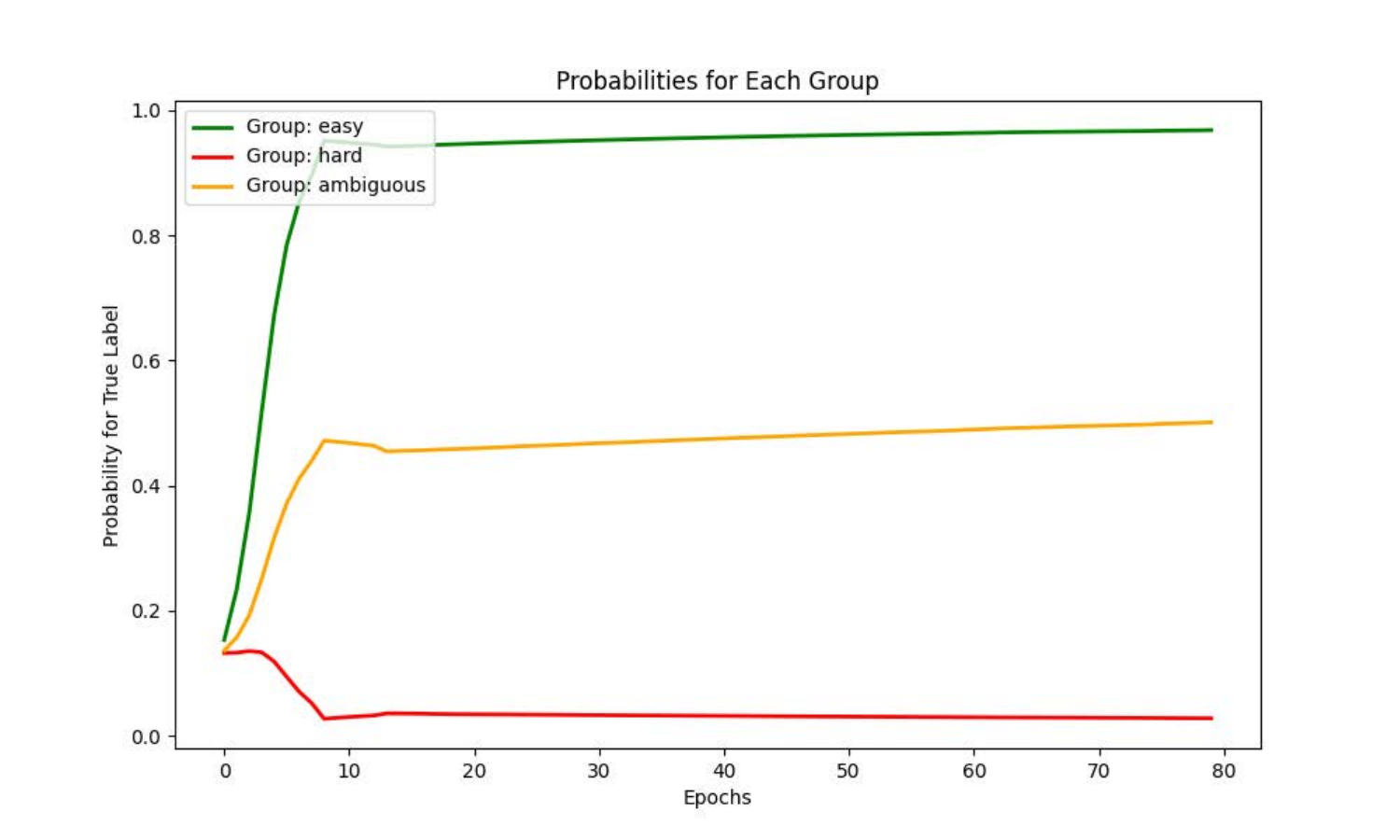}
        \caption{cora graphsage}
    \end{subfigure}
    \hspace{0.01\linewidth}
    \begin{subfigure}[b]{0.48\linewidth}
        \includegraphics[width=\linewidth]{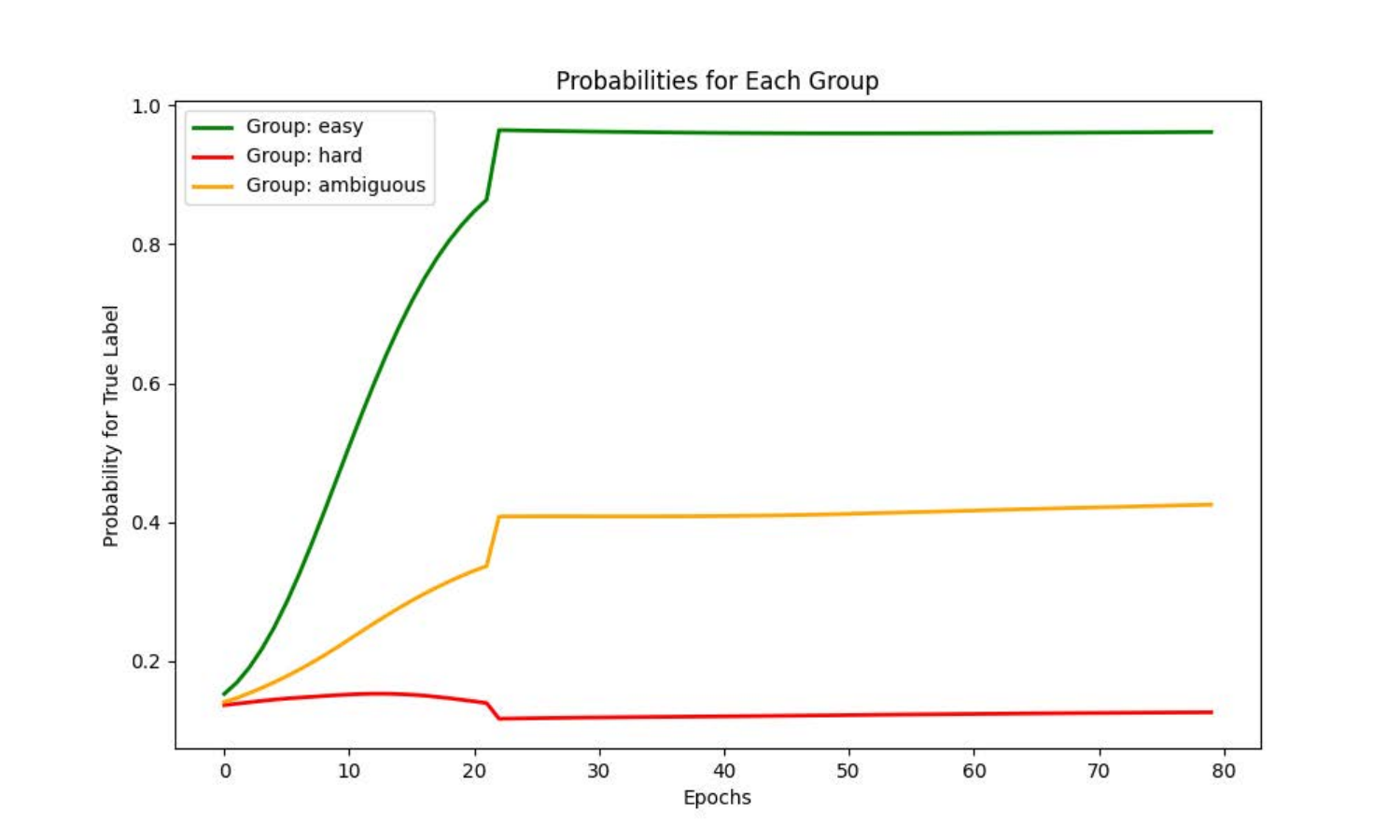}
        \caption{cora mlp}
    \end{subfigure}
    \caption{Training dynamics for \cora dataset using \gcn, \gat, \gs, and \mlp.}
    \label{fig:training_dynamics_cora}
\end{figure}

As shown in Figure \ref{fig:training_dynamics_cora}, across all models, the easy group shows a consistent and rapid increase in predicted probability, quickly converging to high confidence in the correct class. The \mlp, despite lacking neighborhood aggregation, still achieves near-perfect accuracy on these easy nodes, indicating that their correct classification is largely feature-driven.

In contrast, the hard group consistently receives very low predicted probabilities for the true class, regardless of the model architecture. For the \gcn, \gs, and \mlp in particular, the probability for hard nodes stagnates early or even declines slightly during training. This suggests that these nodes are either poorly represented in the feature space or misaligned with their neighborhoods, making them persistently difficult to classify.

The ambiguous group occupies an intermediate position in terms of learning dynamics. The rate and extent of this improvement vary by architecture: \gcn and \gat show relatively stronger performance, while, \gs and \mlp show weaker progression. This indicates that the structures of \gcn and \gat are more effective at incorporating structural cues to disambiguate challenging examples. In contrast, \gs exhibits only modest gains, suggesting limited flexibility in adapting to uncertain or noisy neighborhoods. \mlp, which lacks access to graph structure entirely, demonstrates the slowest improvement, relying solely on node features and showing limited capacity to resolve ambiguity through training alone. These differences underscore the importance of relational inductive bias in improving performance on less clearly defined instances, and position the ambiguous group as a sensitive indicator of a model’s capacity for representation refinement and generalization.

Overall, the training dynamics reveal that while all models are highly effective at fitting easy nodes, they struggle with hard nodes. The ambiguous group provides an informative gradient of difficulty, and the degree to which its performance improves reflects each model's capacity for generalization. These observations motivate further exploration of specialized training strategies or architectures to better handle ambiguous and hard examples in graph-structured data.

\paragraph{\credit}

\begin{figure}[!ht]
    \centering
    \begin{subfigure}[b]{0.48\linewidth}
        \includegraphics[width=\linewidth]{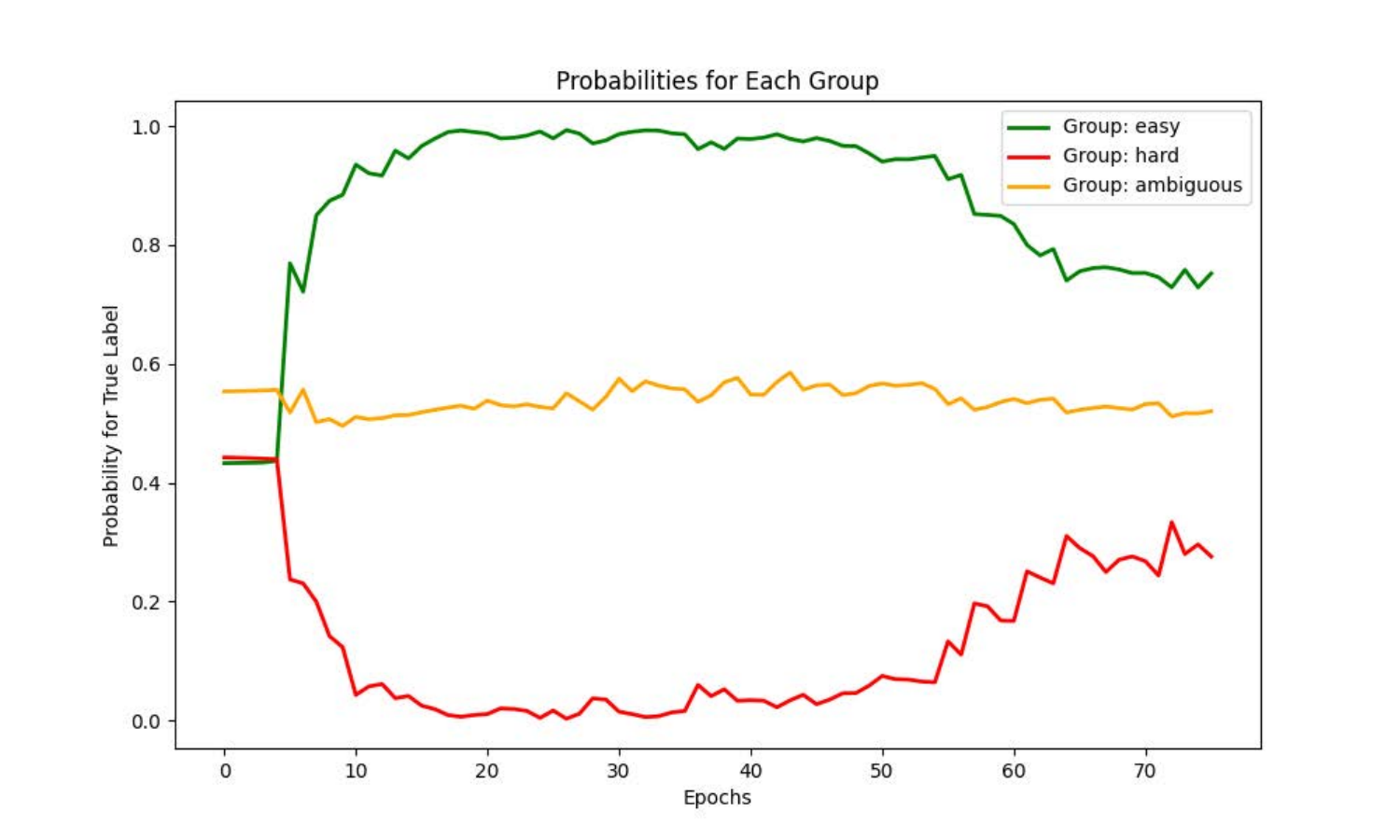}
        \caption{credit gcn}
    \end{subfigure}
    \hspace{0.01\linewidth}
    \begin{subfigure}[b]{0.48\linewidth}
        \includegraphics[width=\linewidth]{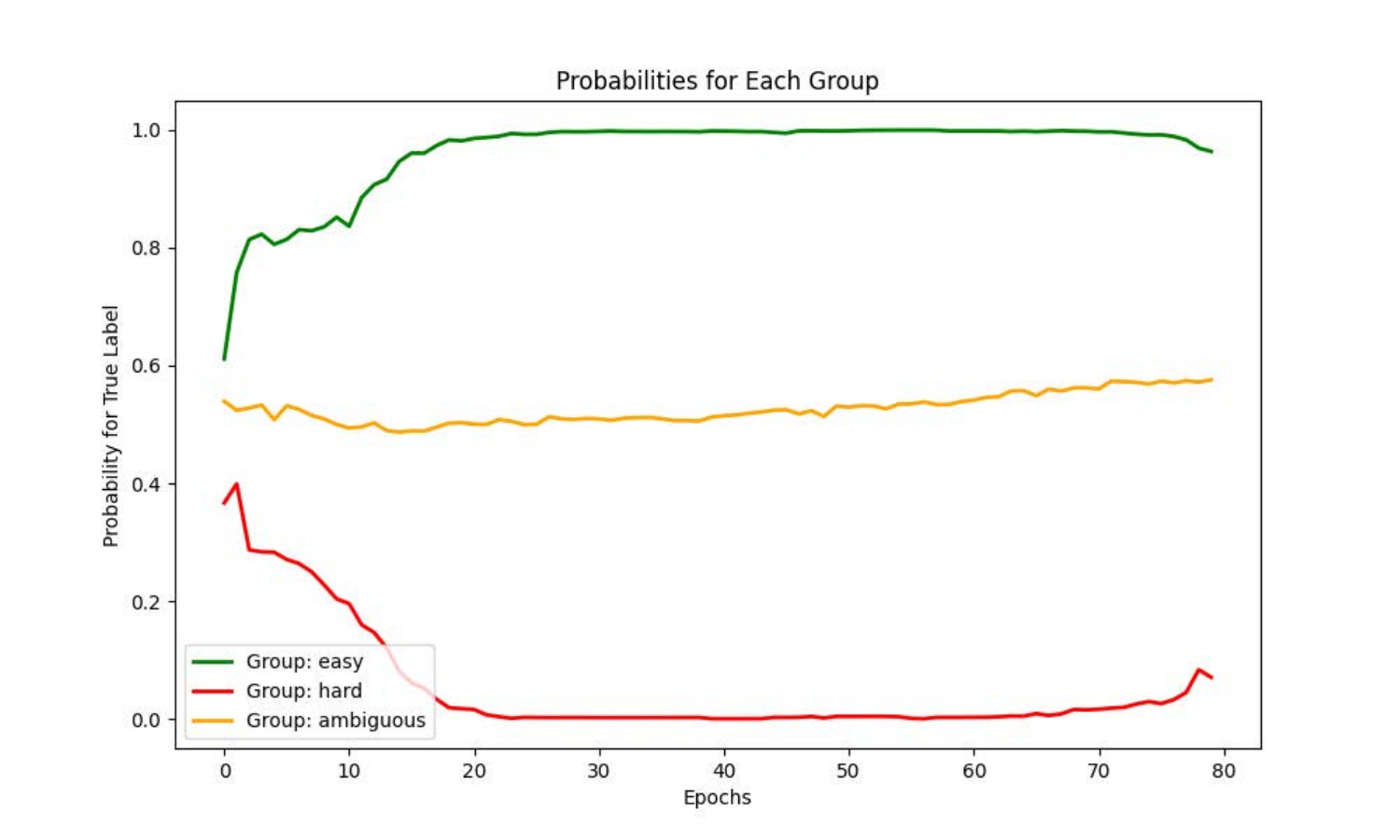}
        \caption{credit gat}
    \end{subfigure}
    \vspace{2mm}
    \begin{subfigure}[b]{0.48\linewidth}
        \includegraphics[width=\linewidth]{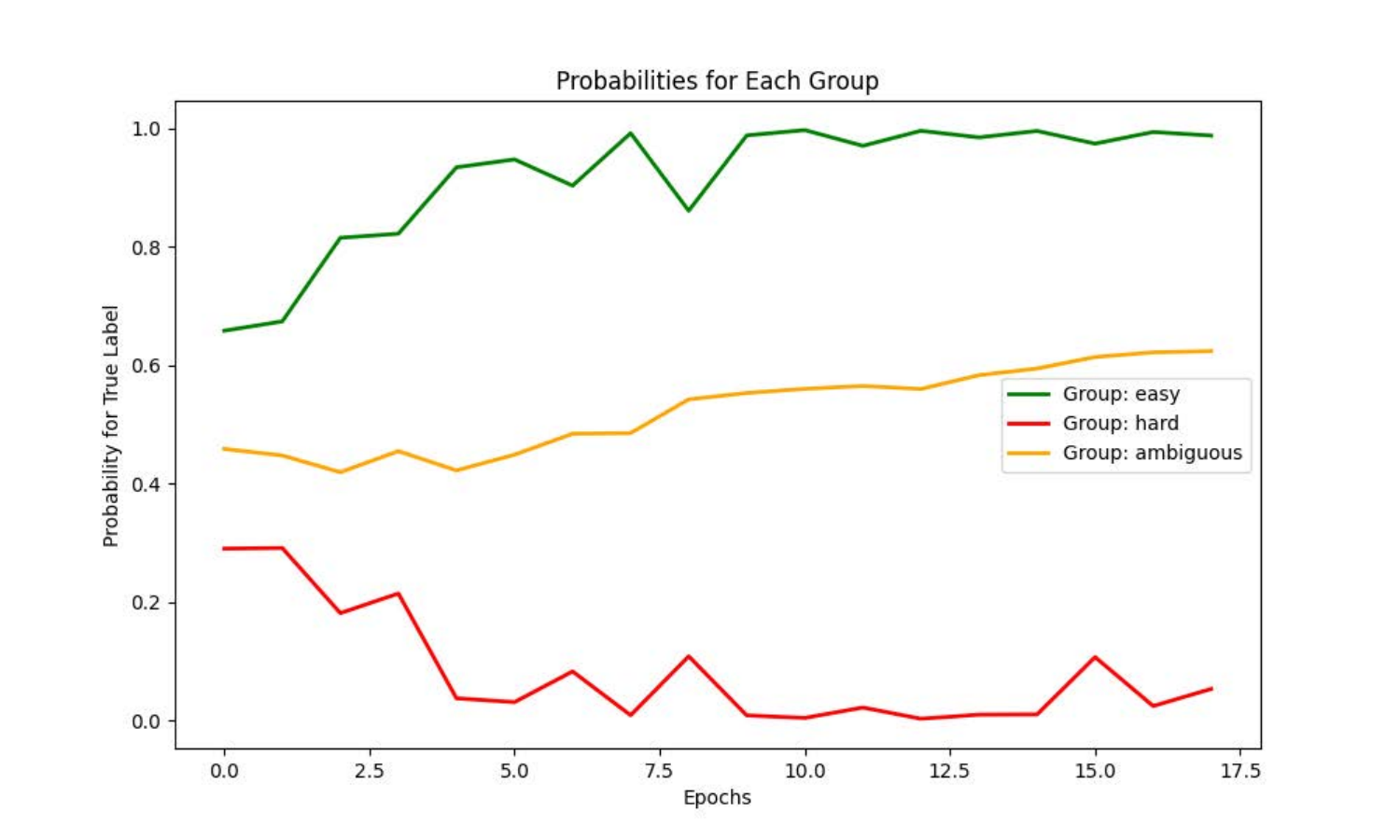}
        \caption{credit graphsage}
    \end{subfigure}
    \hspace{0.01\linewidth}
    \begin{subfigure}[b]{0.48\linewidth}
        \includegraphics[width=\linewidth]{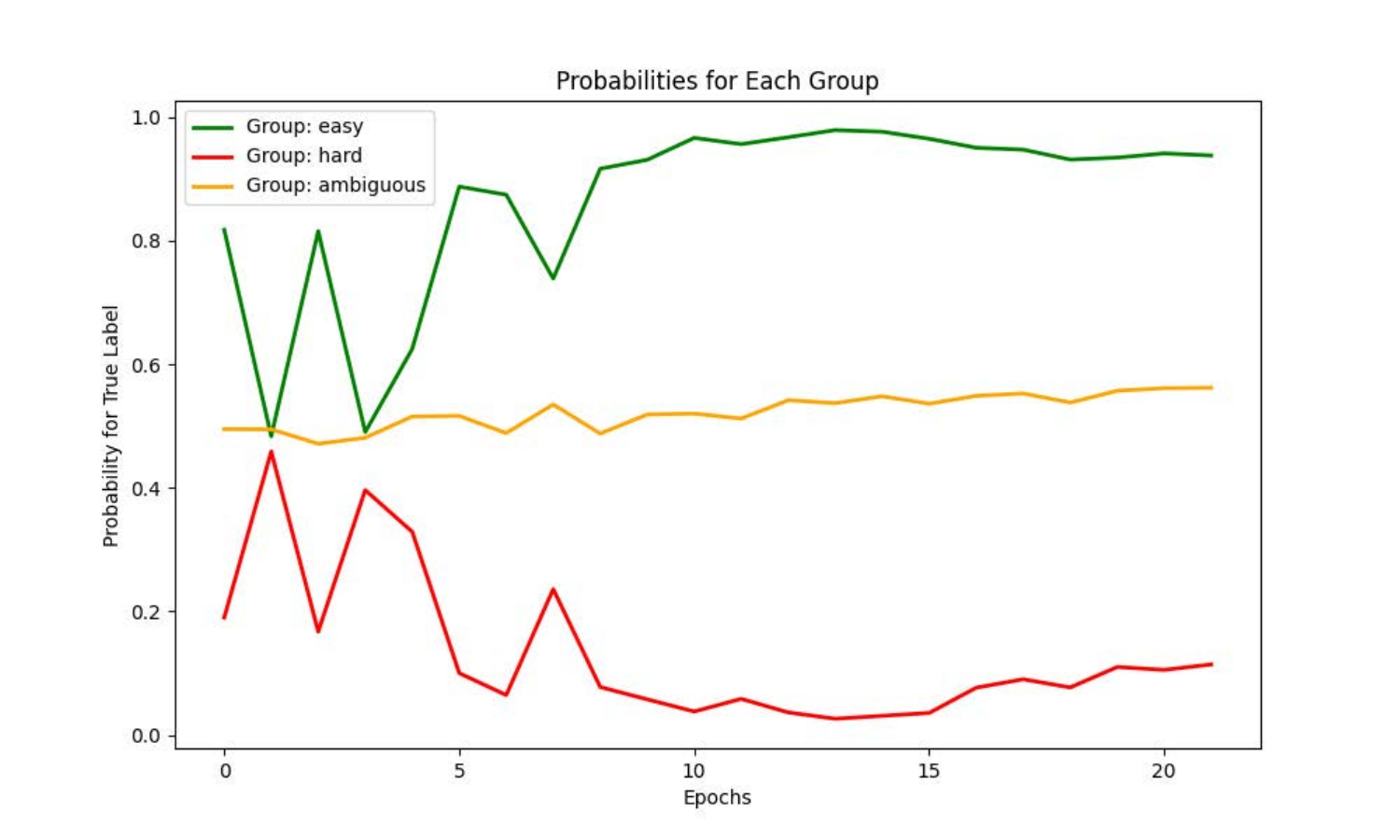}
        \caption{credit mlp}
    \end{subfigure}
    \caption{Training dynamics for credit dataset using \gcn, \gat, \gs, and \mlp.}
    \label{fig:train_dynamics_credit}
\end{figure}

As shown in Figure \ref{fig:train_dynamics_credit}, despite the clear separation between groups, the curves on \credit exhibit greater fluctuation compared to other datasets, indicating a less stable training process. 

The easy group consistently reaches high predicted probabilities across all models, typically plateauing above $0.9$. This suggests that the features associated with these nodes are highly discriminative, enabling models, even those lacking relational structure such as \mlp, to quickly gain confidence in their predictions. 

Interestingly, during the early stages of training, as the confidence of the model in the easy nodes increases, the predicted probabilities for the hard nodes tend to decrease. This pattern reflects an early overfitting tendency, where the model prioritizes easily learnable patterns at the expense of more challenging examples. In the later stages of training, particularly for \gcn and \gat, the predicted probabilities for the hard group begin to rise, however, this improvement comes with a trade-off, as the confidence in the easy group starts to decline. This shift suggests a redistribution of model focus that may reflect capacity limitations or conflicting gradients across subgroups.

The ambiguous group occupies a stable middle ground. For \gcn, \gat, and \mlp the predicted probability for this group remains relatively flat. \gs exhibits a similar trajectory but with slightly more upward movement.

\begin{figure}[h!]
    \centering
    \begin{subfigure}[b]{0.49\linewidth}
        \includegraphics[width=\linewidth]{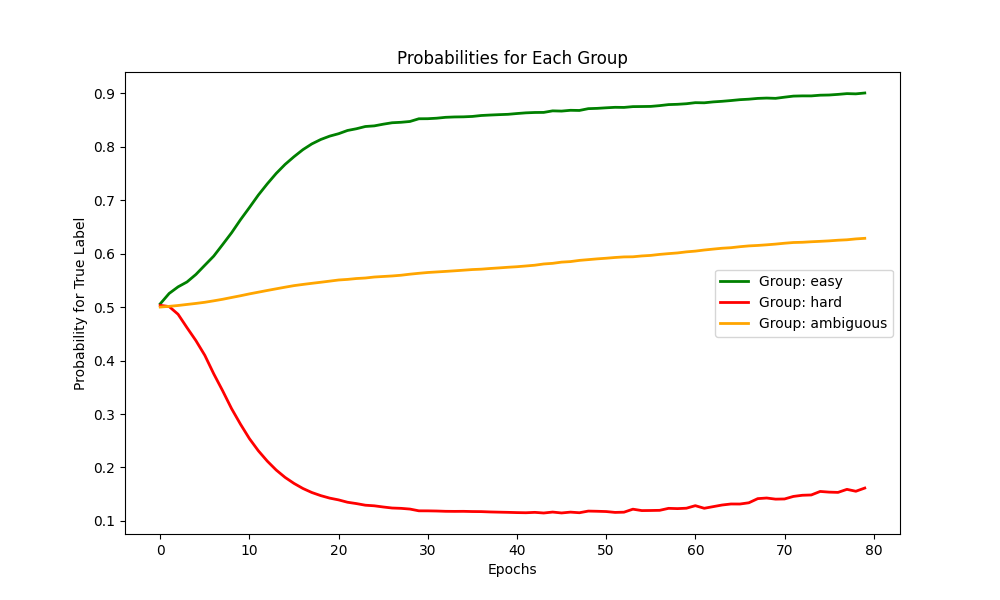}
        \caption{bitcoin gcn}
    \end{subfigure}
    \hspace{0pt}
    \begin{subfigure}[b]{0.49\linewidth}
        \includegraphics[width=\linewidth]{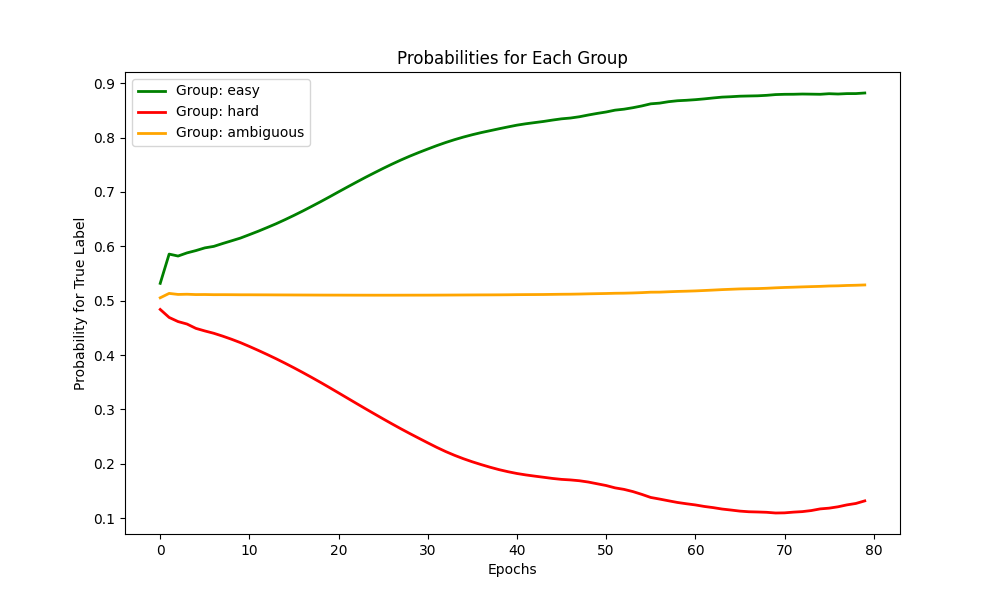}
        \caption{bitcoin gat}
    \end{subfigure}
    %\vspace{2mm}
    \begin{subfigure}[b]{0.49\linewidth}
        \includegraphics[width=\linewidth]{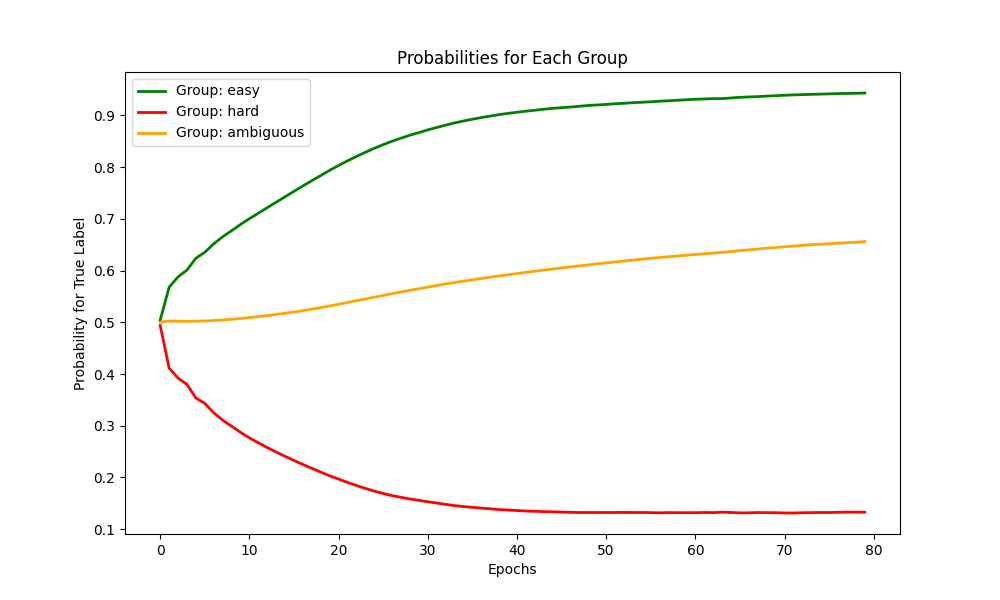}
        \caption{bitcoin graphsage}
    \end{subfigure}
    \hspace{0pt}
    \begin{subfigure}[b]{0.49\linewidth}
        \includegraphics[width=\linewidth]{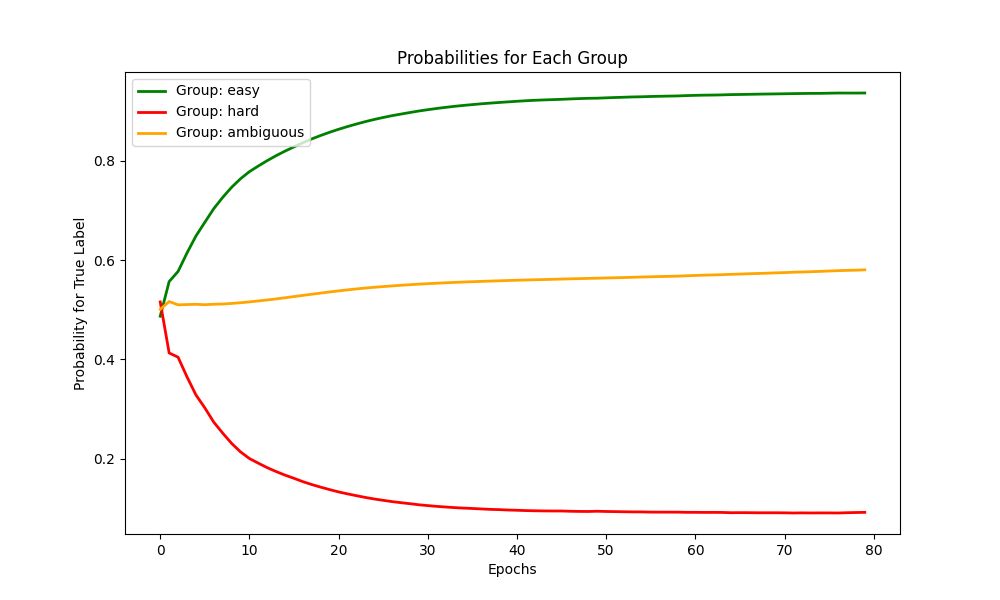}
        \caption{bitcoin mlp}
    \end{subfigure}
    \caption{Training dynamics for Bitcoin dataset using GCN, GAT, GraphSAGE, and MLP.}
    \label{fig:training_dynamics_bitcoin}
    
\end{figure}

\paragraph{\bitcoin} Overall, the categorization of nodes into easy, ambiguous, and hard demonstrates the consistent ability of \mymodel to distinguish between different characteristics of the nodes. As shown in Figure \ref{fig:training_dynamics_bitcoin}, all evaluated models exhibit high confidence and low variance on the easy nodes. In contrast, ambiguous nodes display moderate and gradual improvement, while hard nodes consistently show low confidence and greater variance throughout training.

Both \gcn and \gs rapidly increase their confidence on easy nodes during the early training stages, reaching probabilities near $0.9$, which then remain stable until the end of training. In comparison, ambiguous nodes improve more slowly, converging around $0.6$ by the final epochs. Interestingly, during the early training phase, when confidence on easy nodes is rising, the confidence on hard nodes actually declines from approximately $0.5$ to $0.1$. Notably, \gcn shows a slight late-stage recovery for hard nodes after the $60$-th checkpoint.

\gat follows a similar trend to \gcn in its treatment of node subgroups. Easy nodes are learned quickly, reaching around $0.88$ in probability, while ambiguous nodes exhibit minimal gains over training and converge near their initialization level of $0.51$. Among the models considered, \gat demonstrates the least improvement on ambiguous nodes. For hard nodes, the predicted probability for the true label starts near $0.5$ due to random initialization but steadily drops to approximately $0.1$ as the model focuses on optimizing performance for easier examples.

Interestingly, \mlp performs comparably to other GNN models, albeit with slightly lower confidence on the easy nodes. This is likely because topological information, such as in-degree, out-degree, and edge weights, is incorporated as part of the node input features.

%\todo[inline]{Rwmove the white sapce to make the images in Figure 18 larger}

%%%%%%%%%%%%%%%%%%%%%%%%%%%%%%%%%%%%%%%%%%%%%%%%%%%%%%%%%%%%
\subsection{\fbk Without Flipping Labels}
\label{AppendixSection:NoFlipLabels}
We also investigate the categorization of nodes in the \fbk dataset without applying label flipping. In this setting, the nodes containing randomly replaced entities with label $0$ effectively form a separate class that the \gs model attempts to fit. Table~\ref{tab:fbk_no_flip} presents the categorization statistics generated by \mymodel when training \gs on this modified dataset, which we refer to as FB15-K-Random.

\begin{table}[!ht]
    \centering
    \begin{tabular}{c|c|c|c}
         & easy & ambiguous & hard \\\hline
         &$0.493$  &$0.507$ &$0.00$ \\\hline

    \end{tabular}
    \caption{Categorization percentage assigned by \mymodel when training \gs on the dataset \fbk with $1\%$ randomly chosen nodes injected with random replaced entities.}
    \label{tab:fbk_no_flip}
\end{table}

As shown above, all nodes were categorized into either the easy or ambiguous groups, with none falling into the hard category. Additionally, we observe that the training accuracy is nearly stationary and begins at over $0.9$, which further justifies the need to inject flipped labels to introduce learning difficulty.

\end{document}